\documentclass[10pt,twocolumn,letterpaper]{article}

\usepackage[pagenumbers]{cvpr} %

\usepackage{fontawesome5}%

\usepackage{multirow}
\usepackage{colortbl}

\definecolor{constructionC}{HTML}{7570B3}%
\definecolor{operationC}{HTML}{FC8D62}%
\definecolor{useC}{HTML}{66C2A5}%
\definecolor{constructionFill}{HTML}{E3E2F0}%
\definecolor{operationFill}{HTML}{FEE8E0}%
\definecolor{useFill}{HTML}{E0F3ED}%
\colorlet{taskFill}{black!7}
\definecolor{classBand}{HTML}{E9F2FC}

\usepackage{pifont}
\definecolor{markYes}{HTML}{2E9E44}
\definecolor{markNo}{HTML}{D0342C}
\newcommand{\cmark}{\textcolor{markYes}{\ding{51}}}
\newcommand{\xmark}{\textcolor{markNo}{\ding{55}}}
\newcommand{\hlbox}[2]{{\setlength{\fboxsep}{1pt}\colorbox{#1}{#2}}}

\newcounter{finding}
\newsavebox{\findingbox}
\newenvironment{finding}[1]{%
  \par\smallskip\noindent\colorlet{findingline}{#1}\colorlet{findingfill}{findingline!20}%
  \begin{lrbox}{\findingbox}\begin{minipage}[t]{\dimexpr\linewidth-1.5pt-2\fboxsep\relax}%
  \refstepcounter{finding}\textbf{Finding~\thefinding.}\ }%
  {\end{minipage}\end{lrbox}%
  {\color{findingline}\vrule width 1.5pt}\colorbox{findingfill}{\usebox{\findingbox}}\par\smallskip}

\definecolor{aboveChance}{HTML}{22A05A}
\definecolor{belowChance}{HTML}{D64646}
\newcommand{\acc}[3][]{\ifdim #2pt<#3pt\relax\cellcolor{belowChance!\fpeval{round(min(1,(#3-#2)/#3)*55)}}\else\cellcolor{aboveChance!\fpeval{round(min(1,(#2-#3)/(100-#3))*55)}}\fi{#1#2}}
\newcommand{\accd}[3]{\ifdim #1pt<#2pt\relax\cellcolor{belowChance!\fpeval{round(min(1,(#2-#1)/#2)*55)}}\else\cellcolor{aboveChance!\fpeval{round(min(1,(#1-#2)/(100-#2))*55)}}\fi{#3}}

\newcommand{\fw}{SpatialProbe}        %
\newcommand{\bench}{SpatialProbe-Bench} %

\definecolor{cvprblue}{rgb}{0.21,0.49,0.74}
\usepackage[pagebackref,breaklinks,colorlinks,allcolors=cvprblue]{hyperref}
\hypersetup{pdftitle={SpatialProbe: Diagnosing Internal 3D Representations of Vision-Language Models for Fine-Grained Spatial Reasoning},
  pdfauthor={Jingwen Yang, Yifan Yin, Jianwen Xie, Tianmin Shu}}

\title{\fw{}: Diagnosing Internal 3D Representations of Vision-Language Models for Fine-Grained Spatial Reasoning}

\author{Jingwen Yang$^{1,*}$, Yifan Yin$^{1}$, Jianwen Xie$^{2}$, Tianmin Shu$^{1}$\\[6pt]
$^{1}$Johns Hopkins University \quad $^{2}$Lambda\\[8pt]
\href{https://bi8bo001.github.io/spatialprobe-website/}{\textcolor{black}{\faGlobe}\ \textcolor{cvprblue}{Project Page}}\qquad\quad
\href{https://github.com/Bi8bo001/SpatialProbe}{\textcolor{black}{\faGithub}\ \textcolor{cvprblue}{Code}}\qquad\quad
\href{https://huggingface.co/datasets/bibabooo/SpatialProbe-Bench}{\raisebox{-0.17em}{\includegraphics[height=1.05em]{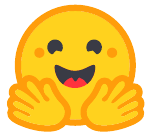}}\ \textcolor{cvprblue}{Data}}
}

\begin{document}
\twocolumn[{%
\renewcommand\twocolumn[1][]{#1}%
\maketitle
\begin{center}
  \captionsetup{type=figure}
  \includegraphics[width=\textwidth]{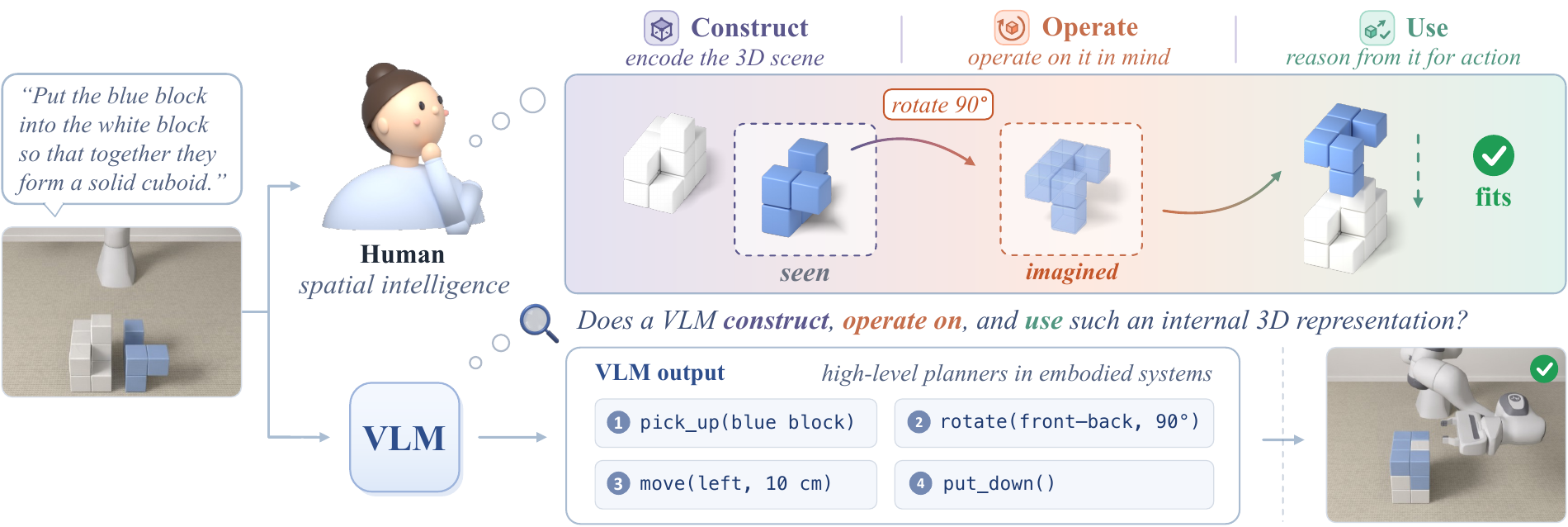}
  \caption{\textbf{Illustration of our research question.} Deciding whether the blue block fits into the white block takes fine-grained spatial reasoning, which requires imagining beyond what is seen, from the position of the block to its orientation and geometry. A human mentally constructs an operable 3D representation of the scene, rotates the block in it, and reasons from the result for further action. We ask whether VLMs, as high-level planners in embodied systems, also \hlbox{constructionFill}{construct}, \hlbox{operationFill}{operate} on, and \hlbox{useFill}{use} internal 3D representations.}
  \label{fig:teaser}
\end{center}
\vspace{1em}
}]
{\renewcommand{\thefootnote}{\fnsymbol{footnote}}%
\footnotetext[1]{Work done during an internship at Johns Hopkins University.}}
\begin{abstract}
Current vision-language models (VLMs) achieve strong spatial understanding of what they see, yet fall far behind humans in fine-grained spatial reasoning, which requires imagining beyond what is seen. Inspired by the cognitive reasoning chain in which humans mentally construct a 3D representation of the scene, operate on it, and reason from it for further action, we ask whether VLMs construct, operate on, and use internal 3D representations. Existing benchmarks score only the final outcome, while representation-level diagnostic studies mainly probe object positions in static scenes, leaving open whether these representations are operable at a fine granularity. To this end, we propose \fw{}, a representation-level diagnostic framework that follows this cognitive reasoning chain to probe a VLM's hidden states in three dimensions (construction, operation, and use) at increasing granularity from position to orientation and geometry. We further build \bench{}, which pairs a diagnostic set that directly evaluates a specific property for fine-grained spatial reasoning in the internal representations with a closed-loop planning set that requires the same reasoning property for success.  Our experiments test and analyze a diverse set of VLMs. We find that the finer the 3D property, the earlier the internal 3D representations of VLMs break, both at an earlier step of this cognitive reasoning chain and at an earlier representation along inference from the input to the answer. These findings highlight the importance of operable internal 3D representations for fine-grained spatial reasoning.  %
\end{abstract}

\begin{table*}[t]
\caption{\textbf{Comparison with prior diagnostic studies of 3D representations in VLMs.} \cmark{} the study tests
this, \xmark{} it does not. Img.: the image embeddings of the object; VObj.: the object's image tokens inside the language model; TObj.: the object's name token in the query; Ans.: the answer token. Our diagnosis is broader and reaches finer 3D properties, across all four representations along inference.}
\label{tab:related}
\centering
\footnotesize
\setlength{\aboverulesep}{0pt}
\setlength{\belowrulesep}{0pt}
\renewcommand{\arraystretch}{1.15}
\setlength{\tabcolsep}{5pt}
\begin{tabular}{@{\hspace{3pt}}lcccccccccc@{\hspace{3pt}}}
\toprule
\multirow{3}{*}{Diagnostic study} & \multicolumn{6}{c}{Evaluated dimension} & \multicolumn{4}{c}{Evaluated representation} \\
 & \multicolumn{3}{c}{\cellcolor{constructionFill}Construction} & \multicolumn{2}{c}{\cellcolor{operationFill}Operation} & \multicolumn{1}{c}{\cellcolor{useFill}Use} & \multicolumn{4}{c}{\cellcolor{classBand}Along inference} \\
 & \textit{Position} & \textit{Orientation} & \textit{Geometry} & \textit{Forward} & \textit{Inverse} & \textit{Edit} & \textit{Img.} & \textit{VObj.} & \textit{TObj.} & \textit{Ans.} \\
\midrule
Hidden in Plain Sight~\cite{hiddenplainsight} & \cmark & \xmark & \cmark & \xmark & \xmark & \xmark & \cmark & \cmark & \xmark & \xmark \\
Spatial IDs~\cite{spatialids} & \cmark & \xmark & \xmark & \xmark & \xmark & \cmark & \xmark & \cmark & \cmark & \xmark \\
MLLM Orientation~\cite{mllmorientation} & \xmark & \cmark & \xmark & \xmark & \cmark & \xmark & \cmark & \xmark & \xmark & \xmark \\
Scene Topology~\cite{scenetopology} & \cmark & \xmark & \xmark & \xmark & \xmark & \xmark & \xmark & \cmark & \xmark & \xmark \\
Why Far Looks Up~\cite{farlooksup} & \cmark & \xmark & \xmark & \xmark & \xmark & \xmark & \xmark & \xmark & \xmark & \cmark \\
RotOutBench~\cite{rotoutbench} & \xmark & \xmark & \xmark & \cmark & \xmark & \xmark & \xmark & \xmark & \xmark & \cmark \\
Knowing Isn't Always Saying~\cite{knowingsaying} & \xmark & \xmark & \xmark & \xmark & \xmark & \cmark & \xmark & \xmark & \cmark & \cmark \\
S-Space~\cite{sspace} & \cmark & \xmark & \xmark & \cmark & \xmark & \cmark & \xmark & \xmark & \cmark & \xmark \\
\midrule
\textbf{\fw{} (ours)} & \cmark & \cmark & \cmark & \cmark & \cmark & \cmark & \cmark & \cmark & \cmark & \cmark \\
\bottomrule
\end{tabular}
\end{table*}
\section{Introduction}
\label{sec:intro}

Spatial intelligence is a fundamental building block of human cognition \cite{spelke2007core}. It is also a central ability for embodied agents to robustly interact with the physical world~\cite{spinbench,vsibench}.
In recent years, vision-language models (VLMs) have achieved strong spatial understanding of what they directly see, such as object counts, distances and spatial relations~\cite{vsibench,omnispatial}. They have also increasingly served as planners in embodied systems~\cite{palme,robobrain,geminirobotics15} and as pretrained backbones for many vision-language-action (VLA) policies~\cite{rt2,openvla,pi0}. In these roles, they are expected not only to understand the 3D scenes they observe but also to imagine the outcomes of potential actions before execution.

However, current VLMs still lag far behind humans on tasks requiring imagination beyond what is seen, such as mental rotation~\cite{spinbench}, mental folding~\cite{stare,spatialvizbench}, and perspective taking~\cite{mindcube,omnispatial}.
Moreover, embodied settings such as manipulation require this ability at a finer granularity, which we call fine-grained spatial reasoning.
For example, inserting the blue block into the white block (\cref{fig:teaser}) requires a planner to go beyond their visible positions and relations.
It must first imagine the blue block's position as it is lifted to avoid collision.
It must then imagine how the block's orientation changes when rotated about different axes by different angles.
Finally, it should reason about both blocks' geometry to decide whether the blue block fits.
Fine-grained spatial reasoning also extends to object parts, such as where a mug's handle will point after a rotation.
Such part-level tasks remain challenging for current robot manipulation models~\cite{partinstruct}.
\begin{figure*}[t]
  \centering
  \includegraphics[width=\textwidth]{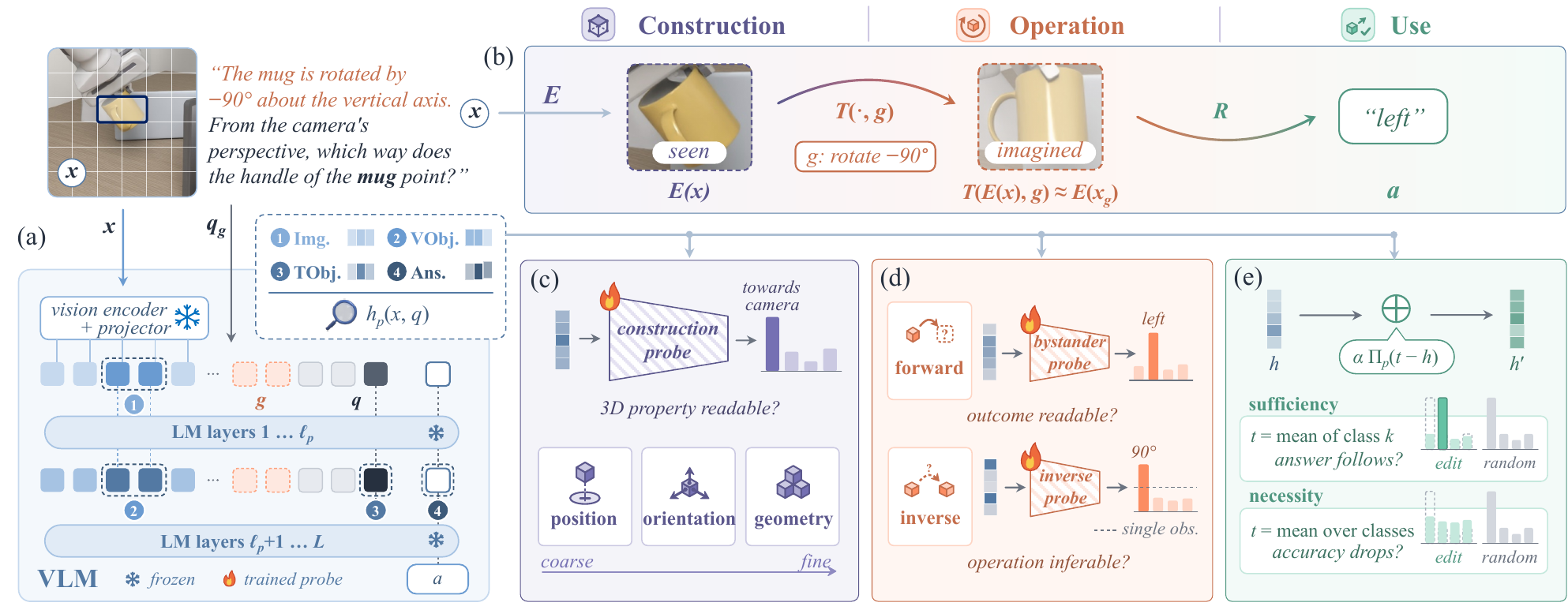}
  \vspace{-10pt}
  \caption{\textbf{Overview of our \fw{} diagnostic framework.} (a) The four internal 3D representations that we read from the VLM along inference, namely the image embeddings (Img.), the object's image tokens (VObj.), the object's name token in the query (TObj.) and the answer token (Ans.). (b) The cognitive reasoning chain, in which a 3D representation of the scene is \hlbox{constructionFill}{constructed}, \hlbox{operationFill}{operated on} and \hlbox{useFill}{used} to produce the answer. (c) to (e) \fw{} diagnoses along the cognitive reasoning chain step by step on each of the four internal 3D representations. Notation follows \cref{sec:framework}.}%
  \label{fig:framework}
  \vspace{-10pt}
\end{figure*}

In contrast, humans can solve such fine-grained spatial reasoning tasks intuitively.
When dealing with the task in \cref{fig:teaser}, a person can form a picture of the 3D scene in mind, rotate the blue block mentally, and imagine whether it fits~\cite{shepard1971mental}.
Inspired by this cognitive reasoning chain of constructing, operating on and using a mental 3D representation, we ask whether VLMs similarly form an operable internal 3D representation of the scene.

Existing work cannot yet answer this question.
First, prior embodied benchmarks score the final outcome~\cite{robospatial,manipbench,embodiedbench}, and their failure analyses show which step fails but not where inside the model the fine-grained spatial reasoning breaks.
Second, as shown in \cref{tab:related}, current representation-level diagnostic studies read the hidden states of VLMs, but their scope is narrow: they mainly probe object positions in static scenes~\cite{hiddenplainsight,spatialids,scenetopology,farlooksup} and test operations only for positions and in-plane orientation~\cite{sspace,rotoutbench,mllmorientation}.
Moreover, each study reads a limited set of representations rather than tracing them along inference.
For example, S-Space~\cite{sspace} only reads object positions at the object's name token, and MLLM Orientation~\cite{mllmorientation} probes in-plane rotations between two views from the image embeddings.
Therefore, it remains open whether internal 3D representations are operable at fine granularity and where along inference they break.

To this end, we design \fw{}, a representation-level diagnostic framework, to answer our research question: \emph{can VLMs construct, operate on, and use internal 3D representations for fine-grained spatial reasoning?}
Specifically, following this cognitive reasoning chain (\cref{fig:framework}), it answers this question in three dimensions by probing VLM hidden states along inference, at increasing granularity from position to orientation and geometry.
\emph{Construction}: can these 3D properties of objects be read from the hidden states?
\emph{Operation}: given an operation, do the internal 3D representations update accordingly, and can the operation be inferred from the observations before and after it?
\emph{Use}: do the answers of the VLM depend on these representations?
Along inference from the input to the answer, our framework reads four internal 3D representations: the image embeddings, the object's image tokens, the object's name token in the query and the answer token.
To apply this framework to an embodied setting, we further build \bench{}, which pairs a diagnostic set for the above three dimensions with a closed-loop downstream planning set of part-level manipulation and geometric assembly, both requiring fine-grained spatial reasoning.
On the planning set, we evaluate VLMs with two harnesses, in which the VLM outputs either one action per turn or code as policies~\cite{codeaspolicies,capx}. We then conduct a failure analysis that finds the decisive error of each failed episode.

Our experiments evaluate diverse VLMs. The results show that the finer the 3D property, the earlier the internal 3D representations of VLMs break, both at an earlier step of the cognitive reasoning chain and at an earlier representation along inference. First, along the cognitive reasoning chain, geometry already breaks at construction and orientation at operation, while position is constructed, operated on and used. Second, along inference, geometry is lost before the object's name token, whereas position and orientation remain readable up to the answer token. The failures on the planning set follow the same pattern.

In summary, our main contributions are as follows:
\begin{itemize}
  \item \fw{}, a representation-level diagnostic framework that probes along inference whether VLMs construct, operate on, and use internal 3D representations for fine-grained spatial reasoning.
  \item \bench{}, a dataset that pairs a diagnostic set for probing the internal 3D representations of VLMs with a closed-loop planning set for analyzing failures in fine-grained spatial reasoning.
  \item A comprehensive analysis of 18 open-source VLMs, showing how and when their internal 3D representations fail to support fine-grained spatial reasoning.
\end{itemize}

\section{Related Work}
\label{sec:related}

\noindent\textbf{Spatial Reasoning in VLMs.}
Spatial reasoning connects what an embodied system sees with how it acts~\cite{omnispatial}.
Current VLMs achieve strong spatial understanding of what they see: tasks on the configuration of a space, such as object counting, distance and relations, are approaching saturation~\cite{omnispatial}.
In contrast, VLMs fall far behind humans when the reasoning requires imagining beyond what is seen, for example in mental rotation and perspective taking~\cite{spinbench,mindcube,omnispatial} or in mental folding and assembly~\cite{stare,spatialvizbench}.
Meanwhile, as VLMs increasingly serve as high-level planners in embodied systems~\cite{palme,robobrain,geminirobotics15}, embodied benchmarks evaluate their spatial reasoning for action~\cite{roborefer,robospatial,manipbench,embodiedbench}.
WM-ABench~\cite{wmabench} further asks whether VLMs have internal world models, through atomic tests of perceiving a scene and predicting the outcome of an action.
However, prior benchmarks only score the final outcome without probing where the spatial reasoning breaks inside the model, while we explicitly diagnose the internal 3D representations for fine-grained spatial reasoning.

\noindent\textbf{3D Representations in VLMs.}
To improve spatial reasoning, recent methods bring 3D representations into the reasoning process of VLMs.
Some train the VLM to generate them while reasoning, as text, latent tokens or images~\cite{mindcube,spatialreasoner,mirage,3dthinker,monet,mvot,wmvlm,electricsheep}; others obtain them from external 3D models and tools~\cite{vlm3r,spatialmllm,vgllm,vadar,spatialpin,apc,mindjourney}.
Meanwhile, representation-level diagnostic studies turn to the VLMs themselves and probe what 3D information their hidden states carry, mainly object positions~\cite{hiddenplainsight,spatialids,scenetopology,farlooksup,sspace} and in-plane orientation~\cite{mllmorientation}.
Some recent works further consider mental operations~\cite{rotoutbench,sspace,mllmorientation} or edit these representations to change the answer~\cite{spatialids,knowingsaying}.
However, as shown in \cref{tab:related}, their evaluation is neither broad nor deep: most studies test one aspect, and mostly at the level of object positions in static scenes.
Moreover, each study reads only one or two representations.
Our framework addresses these gaps through broader diagnosis of internal 3D representations along inference.

\section{Diagnostic Framework}
\label{sec:framework}
Inspired by human spatial intelligence, we propose \fw{}, a novel diagnostic framework for testing whether VLMs can construct, operate on, and use internal 3D representations for fine-grained spatial reasoning (\cref{fig:framework}).

\subsection{Cognitive Reasoning Chain Formulation}
\label{sec:formulation}
Fine-grained spatial reasoning requires imagining the outcome of an operation before seeing it.
We formulate a spatial reasoning question as a pair $(x, q)$, where $x$ is the current observation of a scene and the query $q$ asks about a 3D property of an object in it, such as where the handle of a mug points.
We denote its correct answer by $y(x)$, one of $K$ classes.
We further denote by $g$ an operation on the scene, either given or inferred, such as rotating the mug by 90° about the vertical axis, and by $x_g$ the updated observation of the operated scene.
Imagining its outcome means predicting $y(x_g)$ from $x$ without seeing $x_g$, such as where the handle points after the mug is rotated by 90°.
We cast such imagination through an internal 3D representation as the cognitive reasoning chain (\cref{fig:framework}b):
\begin{equation}
  x \xrightarrow{\;E\;} E(x) \xrightarrow{\;T(\cdot,\, g)\;} T\big(E(x), g\big) \xrightarrow{\;R\;} a ,
  \label{eq:chain}
\end{equation}
where $E$ constructs the representation from the observation, $T$ operates on it with $g$, and $R$ uses the result to produce the answer $a$.
For the operation step, following the notion of equivariance~\cite{lenc2015equivariance,dupont2020equivariant}, we define a representation as operable when it satisfies
\begin{equation}
  T\big(E(x), g\big) \approx E(x_g) ,
  \label{eq:operable}
\end{equation}
which requires operating on the representation of a scene to give the representation of the operated scene.
We test \cref{eq:operable} through the 3D property that $q$ asks about, \ie, $y(x_g)$ should be readable from $T(E(x), g)$.

Our diagnosis follows  along this cognitive reasoning chain step by step to locate where the fine-grained spatial reasoning of VLMs fails internally.
\textbf{Construction} tests whether $y(x)$ can be read directly from $E(x)$.
\textbf{Operation} tests the step from $E(x)$ to $T(E(x), g)$ in two directions, forward and inverse.
The forward direction describes $g$ in words before the query $q$ and tests whether $y(x_g)$ can be read from $T(E(x), g)$.
We denote the resulting query by $q_g$.
The inverse direction tests whether $g$ can be inferred from $x$ and $x_g$.
\textbf{Use} tests whether $R$ produces the answer from $E(x)$ and from $T(E(x), g)$, by editing them and checking whether the answer follows.

\subsection{Internal 3D Representations}
\label{sec:rep_level}
As shown in \cref{fig:framework}a, we diagnose four internal 3D representations of a VLM along inference, from the input to the answer.
A VLM processes $x$ and $q$ as a single sequence of tokens, and $H^{\ell}(x, q)$ denotes their hidden states after layer $\ell$ of its language model, with $\ell = 0$ as its input.
Each representation, indexed by $p$, is given by a set of tokens $S_p$ at one layer $\ell_p$.
On the image side, we read the image embeddings of the object that $q$ asks about, output by the vision encoder and its projector (Img., $\ell_p = 0$), and the object's image tokens inside the language model (VObj.).
On the language side, we read the object's name token in the query (TObj.) and the answer token, the last input token (Ans.).
The representation $h_p(x, q)$, with $p \in \{\text{Img.}, \text{VObj.}, \text{TObj.}, \text{Ans.}\}$ in this order along inference, is the mean of $H^{\ell_p}(x, q)$ over $S_p$.
At each representation $p$, $E(x)$, $E(x_g)$ and $T(E(x), g)$ in \cref{eq:chain,eq:operable} are realized as
\begin{equation}
  \begin{aligned}
    E_p(x) &:= h_p(x, q), \\
    E_p(x_g) &:= h_p(x_g, q), \\
    T_p\big(E_p(x), g\big) &:= h_p(x, q_g).
  \end{aligned}
  \label{eq:realize}
\end{equation}
Since the VLM receives $q_g$ without seeing $x_g$, $T_p$ is the VLM's own operation on $E_p(x)$.

\subsection{Construction}
\label{sec:construction}
To diagnose the first step of the cognitive reasoning chain (\cref{fig:framework}c), construction tests whether the correct answer $y(x)$ can be read directly from $E_p(x)$.
Our queries ask about three 3D properties at increasing granularity through six question types: left/right, near/far and above/below for position, the facing direction of an object or one of its parts for orientation, and whether two polycubes fit together directly, without a rotation, for geometry.

To read $y(x)$ from $E_p(x)$, we train a linear construction probe~\cite{alain2016probes}: $\hat{y}_p(E_p(x)) = \arg\max_k (W_p E_p(x) + b_p)_k$, where $W_p$ and $b_p$ are its weights and bias and $k$ indexes the $K$ answer classes.
Each VLM has one probe per representation and question type, trained on a separate pool of construction questions about objects that the test questions never use.
Each internal 3D representation is read at one layer, the read layer $\ell_p$, where the probes are most accurate on this pool.
\Cref{sec:impl} further describes the probe training and the layer selection, and \cref{sec:suppl_probing} of the supplementary material describes how each representation is read in detail.
We compare the probe's accuracy with the chance level of each question type and with the VLM's own answers.

\subsection{Operation}
\label{sec:operation}
To diagnose the second step of the cognitive reasoning chain (\cref{fig:framework}d), operation tests the update from $E_p(x)$ to $T_p(E_p(x), g)$ in two directions: forward, where $g$ is given, and inverse, where $g$ is inferred.
We consider two kinds of operations: a movement in one of six directions and a rotation about one of three axes.
A movement may change position (left/right and near/far), and a rotation may change position (above/below), orientation and geometry.

\noindent\textbf{Forward.}
The forward direction gives the VLM a forward question $(x, q_g)$ whose query describes in words an operation $g$ that changes its answer, and tests whether $y(x_g)$ can be read from $T_p(E_p(x), g)$.
A probe trained directly on forward questions could learn the effect of $g$ itself.
Instead, we train the probe as in construction, but on questions where $g$ acts on another object rather than the queried one, leaving $y(x)$ unchanged.
We call such an object a bystander object and this probe the bystander probe.
It uses the same read layer $\ell_p$, so that $T_p(E_p(x), g)$ and $E_p(x_g)$ lie in the same space and the layer is not chosen based on operation results.
We score only scenes of question types on which the construction probe at $p$ is above chance and whose $E_p(x)$ and $E_p(x_g)$ it reads correctly, so failure points to the operation rather than failed construction.

On each scene, the forward question forms a question pair with an answer-preserving question, whose query describes an operation that does not change its answer.
We denote by $u_p(g)$ whether the described operation updates $E_p(x)$ on a scored scene, and set $u_p(g) = 1$ when the bystander probe reads both answers of the question pair correctly: $y(x_g)$ from $T_p(E_p(x), g)$ and the unchanged answer from that of the answer-preserving question.
Reading both correctly rules out a readout that changes whenever the queried object is operated on, whatever the operation does.
Even so, a probe may read $y(x_g)$ from the description of an operation alone.
To measure this, the bystander control $g'$ keeps the image and both queries and only replaces the operated object with a bystander object (\eg, another object rather than the mug), so that neither answer changes, and $u_p(g')$ follows the same criterion as $u_p(g)$.
Over the scored scenes, the forward score is
\begin{equation}
  \Delta_{\mathrm{Fwd}} = \Pr\big[u_p(g) = 1\big] - \Pr\big[u_p(g') = 1\big] .
  \label{eq:fwd}
\end{equation}
It is the share of scored scenes on which $E_p(x)$ is updated beyond the rate that the description alone produces.

\noindent\textbf{Inverse.}
The inverse direction tests whether $g$ can be inferred from $x$ and $x_g$.
Given $x$ and then $x_g$, the query of an inverse question asks for one attribute $g^{*}$ of $g$: the direction or distance of a movement, or the axis, sign or angle of a rotation.
An inverse probe $\hat{g}_p$, trained as in construction on a separate pool of inverse questions, reads $g^{*}$ from the representation $h^{\mathrm{inv}}_p$ of this input.
Since one observation alone can reveal some of $g$, the inverse score subtracts the accuracy of a single-observation probe $\hat{g}_1$ on $E_{\text{Img.}}(v)$, where $v \in \{x, x_g\}$ is the observation that reveals more of $g$:
\begin{equation}
  \Delta_{\mathrm{Inv}} = \Pr\big[\hat{g}_p(h^{\mathrm{inv}}_p) = g^{*}\big] - \Pr\big[\hat{g}_1(E_{\text{Img.}}(v)) = g^{*}\big] .
  \label{eq:inv}
\end{equation}

\subsection{Use}
\label{sec:use}
To diagnose the third step of the cognitive reasoning chain (\cref{fig:framework}e), use tests whether $R$ produces the answer from $E_p(x)$ on construction questions and from $T_p(E_p(x), g)$ after a described operation.
We edit these representations and check whether the VLM's answer follows the edit.
We edit only question types on which the construction probe at $p$ is above chance, so that the 3D property can already be read from the representation before the edit.
Edits act on the tokens $S_p$ at one edit layer per representation, and we denote $h$ as their mean hidden state, computed as $h_p$ at the edit layer rather than at the read layer $\ell_p$.

We denote $\mu_{p,k}$ as the mean of $h$ over the construction probe's training questions of class $k$, and $\bar{\mu}_p$ as the mean of the $\mu_{p,k}$ over the $K$ classes.
The class subspace is spanned by the $\mu_{p,k} - \bar{\mu}_p$, and $\Pi_p$ projects onto it.
Following S-Space~\cite{sspace}, an edit moves only the class component $\Pi_p h$ toward a target $t$ with strength $\alpha$: $ h' = h + \alpha\, \Pi_p (t - h)$. It then adds $h' - h$ to the hidden state of every token in $S_p$; we denote the VLM's answer after the edit by $a(h')$.

\noindent\textbf{Sufficiency.}
To test whether the class component of $h$ suffices to set the answer, we edit it toward a class $k$ different from both the probe's readout and the VLM's answer, using $t = \mu_{p,k}$ and $\alpha > 1$.
The sufficiency score is the increase in $\Pr[a(h') = k]$ relative to a random edit of the same norm orthogonal to the class subspace.

\noindent\textbf{Necessity.}
To test whether the answer requires the class component of $h$, we replace it with that of the mean over classes ($t = \bar{\mu}_p$, $\alpha = 1$).
The necessity score is how much more accuracy drops after this edit than after a random edit of the same norm orthogonal to the class subspace.

\section{Experimental Setup}
\label{sec:setup}

\subsection{\bench{}}
\label{sec:operep}
To diagnose fine-grained spatial reasoning with \fw{} along the cognitive reasoning chain, we need controlled factors and exact 3D ground truth, both to probe the internal 3D representations and to analyze failures in closed-loop planning.
Therefore, we build \bench{}, a synthetic tabletop manipulation dataset in Isaac Sim~\cite{isaacsim}, based on the PartGym environment of PartInstruct~\cite{partinstruct}, with a Franka arm, objects from PartNet-Mobility~\cite{sapien} and custom polycubes (details and examples in \cref{sec:suppl_operep} of the supplementary material).
Its diagnostic set tests each step of the chain, and its closed-loop planning set requires the same fine-grained spatial reasoning to succeed.

\begin{figure*}[t]
  \centering
  \includegraphics[width=\textwidth]{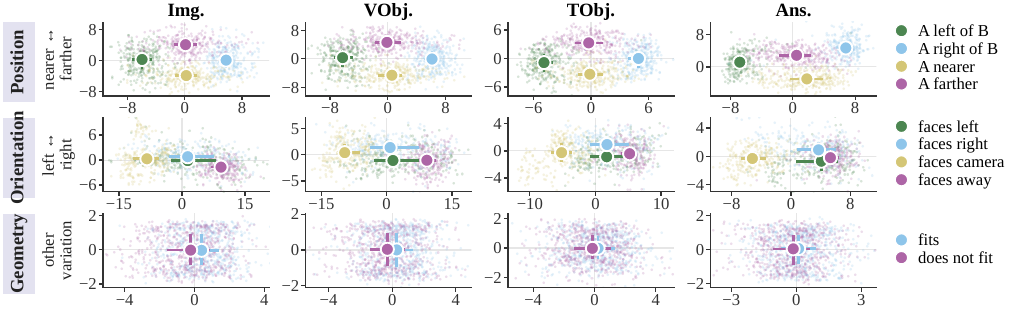}
  \vspace{-10pt}
  \caption{\textbf{Class separation at \hlbox{constructionFill}{construction} along inference.} Hidden states of Qwen2.5-VL-7B at the four representations, projected onto two discriminant axes fitted on other objects. Class separation weakens from position to orientation and geometry, whose two classes largely overlap at all four representations. All 18 VLMs are in \cref{sec:suppl_probing} of the supplementary material. }
  \label{fig:planes}
  \vspace{-10pt}
\end{figure*}

\noindent\textbf{Diagnostic set.}
Each diagnostic question pairs a single-image observation $x$ of a multi-object tabletop scene with a multiple-choice query $q$.
The scenes span several rooms and tabletops, 15 camera viewpoints and 256 objects of nine categories, with the queried object on the table or held by the arm.
Objects and polycube shapes are split between the test questions and the probes' training pool (\cref{sec:construction}).
Viewpoint-dependent questions use both camera-centric and robot-centric reference frames.
For operation, we cover movements in six directions over several distances and rotations about three axes by different angles.
The set holds 7,556 questions, and within each question type, we balance answers across scene factors to avoid spurious correlations.

\noindent\textbf{Planning set.}
The planning set contains two task types requiring fine-grained spatial reasoning, seen from one fixed camera: Part-level Manipulation (114 tasks) and Geometric Assembly (120 tasks).
Part-level Manipulation follows PartInstruct~\cite{partinstruct} in placing and orienting objects down to the part level.
Geometric Assembly asks the robot to infer whether two polycubes are geometrically compatible and to plan accordingly.
Both task types share the same action schema of picking up, moving, rotating and putting down objects.
An episode succeeds when its final state satisfies the goal, and each task is confirmed to be feasible by executing its ground-truth plan successfully in repeated simulation runs.
We evaluate VLMs with two harnesses: Stepwise, in which the VLM outputs one action per turn, and CaP-Agent0~\cite{capx}, in which it outputs code as policies~\cite{codeaspolicies}.
Both execute every action in the physics simulation with motion planned by cuRobo~\cite{curobo}, as detailed in \cref{sec:suppl_task} of the supplementary material.
For the failure analysis, we further categorize the decisive error of each failed episode, the first error that is never recovered, into fine-grained types by the action or 3D property involved, as in the diagnostic set.

\subsection{Models}
\label{sec:models}
We diagnose 18 open-source VLMs across seven series, ranging from 3B to 32B parameters: Qwen2.5-VL~\cite{qwen25vl} (3B, 7B, 32B), Qwen3-VL~\cite{qwen3vl} (4B, 8B, 32B), Qwen3.5~\cite{qwen35} (4B, 9B, 27B), InternVL3.5~\cite{internvl35} (4B, 8B, 14B), Molmo2~\cite{molmo2} (4B, 8B, O-7B), Gemma~3~\cite{gemma3} (4B, 27B) and LLaVA-1.5~\cite{llava15} (7B).
We additionally evaluate the closed-source GPT-6 Astra~\cite{gpt6astra} and Gemini 3.1 Pro~\cite{gemini31pro} on the planning set.
The open-source VLMs run all 234 tasks with Stepwise and 48 of them with CaP-Agent0, while GPT-6 Astra and Gemini 3.1 Pro run these 48 with both harnesses.

\subsection{Implementation Details}
\label{sec:impl}
\noindent\textbf{Probes and edits.}
Each probe is a multinomial logistic regression with an L2 penalty chosen by object-grouped five-fold cross-validation.
Read layers are selected among 12 evenly spaced candidate decoder layers by the construction probes' cross-validated accuracy, and edit layers are either fixed or selected by how much edits change the answers on development scenes, 30\% of all scenes.
\cref{sec:suppl_probing} of the supplementary material gives further details.

\noindent\textbf{Answers and statistics.}
We obtain the VLM's answer by greedy decoding without an edit and by selecting the option with the highest first-token probability after an edit.
The normalized accuracy rescales accuracy so that chance, $1/K$, becomes 0 and a perfect readout 100.
For each VLM, each measure is tested against its control with a one-sided exact test, binomial for construction and paired McNemar~\cite{mcnemar1947} otherwise, under Benjamini-Hochberg correction~\cite{benjamini1995fdr} at 5\% within its dimension.
We report the mean margin above the control over the VLMs, the number of VLMs whose effect is significant, and 95\% bootstrap intervals over scenes.

\section{Results and Analysis}
\label{sec:diagnosis}
Following the cognitive reasoning chain, we locate where the fine-grained spatial reasoning of VLMs fails internally.
We find that the finer the 3D property, the earlier the internal 3D representations of VLMs break, both at an earlier step of the chain (\cref{sec:res_construction,sec:res_operation,sec:res_use}) and at an earlier representation along inference (\cref{sec:res_along}), and the failures on the planning set follow the same pattern (\cref{sec:task_results}).

\subsection{Which 3D Properties Do VLMs Construct?}
\label{sec:res_construction}
\vspace{-10pt}
\begin{finding}{constructionC}
Position and orientation are constructed at all four representations, more accurately than the VLMs answer, whereas geometry already breaks at construction.
\end{finding}

Over the 18 VLMs, the construction probe reads position and orientation above chance at all four representations, up to 48 points more accurately than the VLMs answer, whereas it barely reads geometry.
\Cref{fig:planes} shows this pattern with Qwen2.5-VL-7B as an example: class separation becomes progressively weaker from position to orientation and geometry, with the two classes of geometry overlapping at all four representations.

\noindent\textbf{Position.}
Position is read 35--50 points above chance at every representation, but the answers are as accurate as this readout only on left/right.
Furthermore, the answers lean on 2D image cues rather than on the 3D relation that the representations encode.
As above/below questions show, they are right on 90.0\% of the questions where the cube that appears higher in the image is also above, but on only 34.8\% where it is not, while the readout stays at 81--96\%.

\noindent\textbf{Orientation.}
The orientation of an object part is read less accurately than that of the whole object (30--39 versus 67--82 in normalized accuracy).
Class separation also depends on the facing direction.
Along discriminant axes fitted on other objects, the classes facing toward and away from the camera are separated by 11--21 within-class standard deviations, compared with only 1--2 for those facing left and right (medians over VLMs).

\noindent\textbf{Geometry.}
Geometry is read only slightly above chance, and only at the image side (9.7 and 9.8 points at Img.\ and VObj.).
The little that is read reflects whether the polycube is rotated, not its shape.
For polycubes that do not fit, the probe at VObj.\ correctly predicts ``does not fit'' in 70.8\% of the cases with a rotation mismatch, compared with only 45.3\% of the cases with a shape mismatch.
An MLP probe reads geometry no better (\cref{sec:suppl_probing} of the supplementary material), and the answers stay at chance.

\subsection{Which 3D Properties Do VLMs Operate On?}
\label{sec:res_operation}
\vspace{-10pt}
\begin{finding}{operationC}
A described movement updates position from the object's name token onward, whereas a described rotation updates orientation at none of the representations and only marks the object as operated on.
\end{finding}

\begin{figure}[t]
  \centering
  \includegraphics[width=\columnwidth]{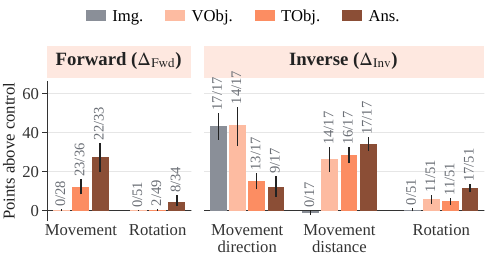}
  \caption{\textbf{\hlbox{operationFill}{Operation} along inference.} Forward (\cref{eq:fwd}) and inverse (\cref{eq:inv}) scores, averaged over the VLMs with 95\% intervals; numbers: significant / measured pairs of VLM and question type. A described movement updates position from TObj.\ onward, whereas a described rotation updates no representation. }
  \label{fig:operation}
  \vspace{-10pt}
\end{figure}

\noindent\textbf{Forward.}
Since the vision encoder never sees the description of $g$, we test the forward direction from VObj.\ onward and place the description before the image for VObj.
A described movement reaches a $\Delta_{\mathrm{Fwd}}$ of 12.2 points at TObj.\ and 27.6 points at Ans.\ (\cref{fig:operation}, left), but only $+0.3$ at VObj.
In contrast, a described rotation updates no representation: its $\Delta_{\mathrm{Fwd}}$ is $+0.6$ at TObj.\ and $+4.7$ at Ans., and stays near zero at every other candidate layer.

\noindent\textbf{Inverse.}
Which way an object moved is read from the image embeddings of the two observations, 43.3 points above the single-observation probe and significantly in every VLM.
However, how far it moved is read only inside the language model, 26.6--34.3 points above that probe from VObj.\ onward (\cref{fig:operation}, right).
For a rotation, the axis and sign are hardly read, and the angle is read only for the smallest rotations: at TObj., the inverse score for the angle is 39.3 points at 30\textdegree{} but at most 3.5 at larger angles.

\noindent\textbf{How a rotation breaks.}
At TObj., a described rotation shifts the representation along one direction shared across questions.
This is the same direction as for a movement that keeps the answer (cosine 0.998, in all 18 VLMs), so the rotation marks the object as operated on rather than rotated.
Even when the query states the orientation before the rotation, the answers treat a rotation as a flip to the opposite class: their accuracy changes by $-3.3$ points on rotations by 90\textdegree{} but rises by 50.7 points on rotations by 180\textdegree{}, where the opposite class is correct (\cref{sec:suppl_answer_level} of the supplementary material).

\subsection{Which 3D Properties Do VLMs Use?}
\label{sec:res_use}
\vspace{-10pt}
\begin{finding}{useC}
The answers use position from the image side on questions with and without an operation, but use orientation only on questions without an operation.
\end{finding}

\begin{figure}[t]
  \centering
  \includegraphics[width=\columnwidth]{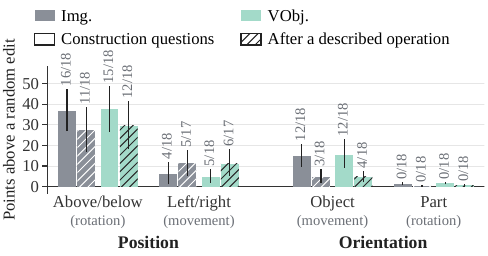}
  \caption{\textbf{\hlbox{useFill}{Use} at the image side.} Sufficiency scores on construction questions (solid) and after a described operation (hatched), averaged over the 18 VLMs with 95\% intervals; numbers: significant / measured VLMs. After an operation, the answers still use position but no longer orientation. }
  \label{fig:use}
\end{figure}

On construction questions (\cref{fig:use}, solid), the sufficiency score at Img.\ and VObj.\ is 36.9 and 37.5 points for above/below and 14.8 and 15.6 for object orientation.
For the orientation of an object part, it is only $+1.3$ to $+1.9$, indicating an earlier break at finer granularity.
Left/right and near/far are instead used at TObj.\ ($+14.8$ and $+24.4$).
After a described operation (\cref{fig:use}, hatched), the image side still drives position answers by 27.4 points for above/below and 11.3 for left/right, but no longer orientation answers: for object orientation, the sufficiency score falls from 14.8 to 4.9 points, declining in all 18 VLMs.
After an operation, the answers no longer need the image side: the necessity score is positive only on construction questions, up to 6.4 points.
Edits at the answer token, in contrast, set the answers in every condition (sufficiency score 12--50 points), so the break lies before the answer token.

Therefore, along the cognitive reasoning chain, the finer the 3D property, the earlier it breaks: geometry already at construction and orientation at operation, while position is constructed, operated on and used.

\subsection{Where Along Inference Does Each 3D Property Break?}
\label{sec:res_along}
\vspace{-10pt}
\begin{finding}{black!60}
Along inference, geometry is lost before the object's name token, whereas position and orientation remain readable up to the answer token, yet the answers express only part of them.
\end{finding}

\begin{figure}[t]
  \centering
  \includegraphics[width=\columnwidth]{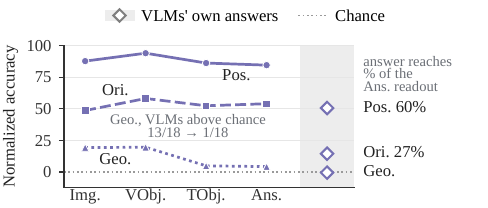}
  \caption{\textbf{\hlbox{constructionFill}{Construction} along inference.} Normalized accuracy of the construction probe, averaged over the question types and the 18 VLMs; diamonds: the VLMs' own answers, with their share of the Ans.\ readout. Pos., Ori.\ and Geo.: position, orientation and geometry. The finer the 3D property, the earlier it breaks. }
  \label{fig:along}
\end{figure}

In normalized accuracy (\cref{fig:along}), position is read at 84--94 from Img.\ to Ans., and orientation at 49--58.
The answers reach only 51 for position and 15 for orientation, which is 60\% and 27\% of the Ans.\ readout.
Geometry falls from 19 and 20 at the image side to 5 and 4 at TObj.\ and Ans., and is answered at 0.
Thus, along inference as well, the finer the 3D property, the earlier it breaks.

\subsection{Do the Failures on the Planning Set Follow the Same Pattern?}
\label{sec:task_results}
\vspace{-10pt}
\begin{finding}{black!60}
The failures on the planning set follow the same pattern as on the diagnostic set: rotation is again the weaker operation, and its outcome cannot be read from the hidden states at the decisive errors.
\end{finding}

\begin{figure}[t]
  \centering
  \includegraphics[width=\columnwidth]{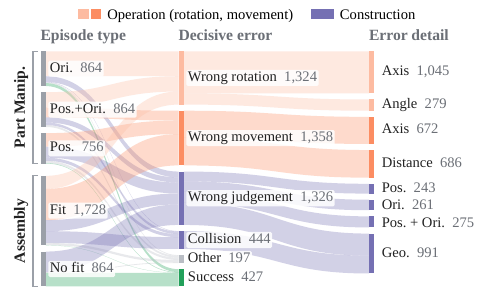}
  \caption{\textbf{Failure analysis on the planning set.} Episodes of the open-source VLMs with both harnesses, from the episode type to the decisive error and its detail. Pos., Ori.\ and Geo.: position, orientation and geometry. }
  \label{fig:task_failures}
\end{figure}
Closed-loop planning requires the same fine-grained spatial reasoning, so we ask whether its failures follow the same pattern.
On the episodes that all VLMs run, the open-source VLMs succeed in only 9.4\% with Stepwise and 4.7\% with CaP-Agent0 on average, and at most 22.9\% (Qwen3.5-27B), whereas GPT-6 Astra and Gemini 3.1 Pro succeed in 54.2\% and 25.0\% with both harnesses (\cref{sec:suppl_task} of the supplementary material).
Of the failed episodes of the open-source VLMs, 28.5\% end with a wrong rotation, 29.2\% with a wrong movement, 28.5\% with a wrong judgement and 9.6\% with a collision (\cref{fig:task_failures}).
As on the diagnostic set, rotation is the weaker operation: in 17 of 18 VLMs, movement actions are right more often than rotation actions, just as the forward score is higher for movement in 16 of 18.
We further probe the hidden states of the Stepwise turns.
At the decisive errors, the state before the action is read at least as well as at matched correct turns (42.3\% versus 32.6\%).
The outcome of a rotation, however, is read no better than from the action text or that state alone in any of the 18 VLMs, as for a described rotation.

\section{Conclusion}
\label{sec:discussion}

To answer whether VLMs can construct, operate on, and use internal 3D representations for fine-grained spatial reasoning, we presented \fw{}, a diagnostic framework along the cognitive reasoning chain, and \bench{}, a dataset with diagnostic and closed-loop planning sets.
We find that the finer the 3D property, the earlier the internal 3D representations break, both at an earlier step of this chain and at an earlier representation along inference, and the failures on the planning set follow the same pattern.

\noindent\textbf{Limitations and future work.}
Our probes are linear and our scenes are synthetic. Future work can extend \bench{} to real scenes and explore training for operable 3D representations in embodied planning.

{
    \small
    \bibliographystyle{ieeenat_fullname}
    \bibliography{main}
}

\clearpage
\maketitlesupplementary
\appendix
\raggedbottom%
\makeatletter\@secpenalty=0\makeatother%

\noindent\textbf{Contents}\par\smallskip
{\small\newcommand{\tocline}[1]{\noindent\makebox[1.6em][l]{\ref{#1}}\nameref{#1}\dotfill\pageref{#1}\par}
\tocline{sec:suppl_notation}
\tocline{sec:suppl_operep}
\tocline{sec:suppl_taxonomy}
\tocline{sec:suppl_diag_results}
\tocline{sec:suppl_probing}
\tocline{sec:suppl_answer_level}
\tocline{sec:suppl_cases}
\tocline{sec:suppl_task}
\smallskip\noindent Citations refer to the references of the main paper.\par}

\medskip\noindent\textbf{Key results}\par
{\small
\noindent\hangindent=1.2em\hangafter=1\makebox[1.2em][l]{1.}The forward score of a described rotation stays near zero at every candidate layer (\cref{fig:depth}), and MLP probes reproduce the linear-probe findings (\cref{fig:mlp}).\par
\noindent\hangindent=1.2em\hangafter=1\makebox[1.2em][l]{2.}Class separation weakens from position to orientation and geometry in 3D visualization and in all 18 VLMs (\cref{fig:suppl_planes3d,fig:suppl_geo_pos,fig:suppl_geo_obj,fig:suppl_geo_geo}).\par
\noindent\hangindent=1.2em\hangafter=1\makebox[1.2em][l]{3.}The answers lean on 2D image cues rather than on the 3D relation that the representations encode, as above/below questions show (\cref{fig:cue}).\par
\noindent\hangindent=1.2em\hangafter=1\makebox[1.2em][l]{4.}Even when the query states the initial orientation, the answers treat a rotation as a flip to the opposite orientation (\cref{fig:flip}).\par
\noindent\hangindent=1.2em\hangafter=1\makebox[1.2em][l]{5.}The open-source VLMs mostly fail the planning set (\cref{fig:suppl_task_success}), and as on the diagnostic set, the outcome of a rotation cannot be read from the hidden states at the decisive errors (\cref{sec:suppl_failures}).\par
}

\section{Notation}
\label{sec:suppl_notation}
\Cref{tab:notation} defines the abbreviations used in the tables and figures.

\begin{table}[!t]
\caption{\textbf{Abbreviations used in the tables and figures.}}
\label{tab:notation}
\centering
\footnotesize
\setlength{\tabcolsep}{3pt}
\setlength{\aboverulesep}{0pt}
\setlength{\belowrulesep}{0pt}
\renewcommand{\arraystretch}{1.15}
\newcommand{\group}[2]{\rowcolor{#1}\multicolumn{2}{@{\hspace{3pt}}l}{\textit{#2}}\\}
\begin{tabular}{@{\hspace{3pt}}l>{\raggedright\arraybackslash}p{0.63\linewidth}@{\hspace{3pt}}}
\toprule
Abbreviation & Meaning \\
\midrule
\group{constructionFill}{Construction}
Pos. & position: left or right, near or far, and above or below \\
Ori. & orientation: the direction of an object's front or of one of its parts \\
Geo. & geometry: whether two polycubes fit together directly, without a rotation \\
\group{operationFill}{Operation}
Rot., Mov. & rotation, movement \\
\group{classBand}{Representations along inference}
Img. & the image embeddings of the object \\
VObj. & the object's image tokens inside the language model \\
TObj. & the object's name token in the query \\
Ans. & the answer token \\
\group{taskFill}{Planning set}
Part Manip. & Part-level Manipulation \\
Assembly & Geometric Assembly \\
Pos., Ori., Pos.+Ori. & Part Manip.\ episode types, named by the 3D property in the goal: position (place the object next to a reference object), orientation (point a part in a direction) or both \\
Fit, No fit & Assembly episodes whose two polycubes fit together after one rotation, or do not \\
\bottomrule
\end{tabular}
\end{table}
\section{\bench{}}
\label{sec:suppl_operep}

The diagnostic set holds 7,556 questions, of which 1,062 are construction questions and 6,494 are operation questions.
The use tests edit the representations of 1,454 questions.
The planning set holds 114 Part-level Manipulation and 120 Geometric Assembly tasks.

\subsection{Scenes and operations}
The scenes vary over four rooms seen from four sides each, eight tabletops and 15 camera viewpoints at three elevations and five azimuths.
Every image has $1024 \times 1024$ pixels.
They hold 256 PartNet-Mobility objects of nine categories and custom polycubes of 48 shapes.
Geometric Assembly uses 84 other polycube shapes.
In geometry questions and in Geometric Assembly, two polycubes may fit together to form a solid cuboid.
An operation moves an object by 10 to 40\,cm in one of six directions or rotates it by $\pm$30°, $\pm$60°, $\pm$90° or 180° about one of three axes.

\subsection{Examples}
\begin{figure*}[t]
  \centering
  \includegraphics[width=\textwidth]{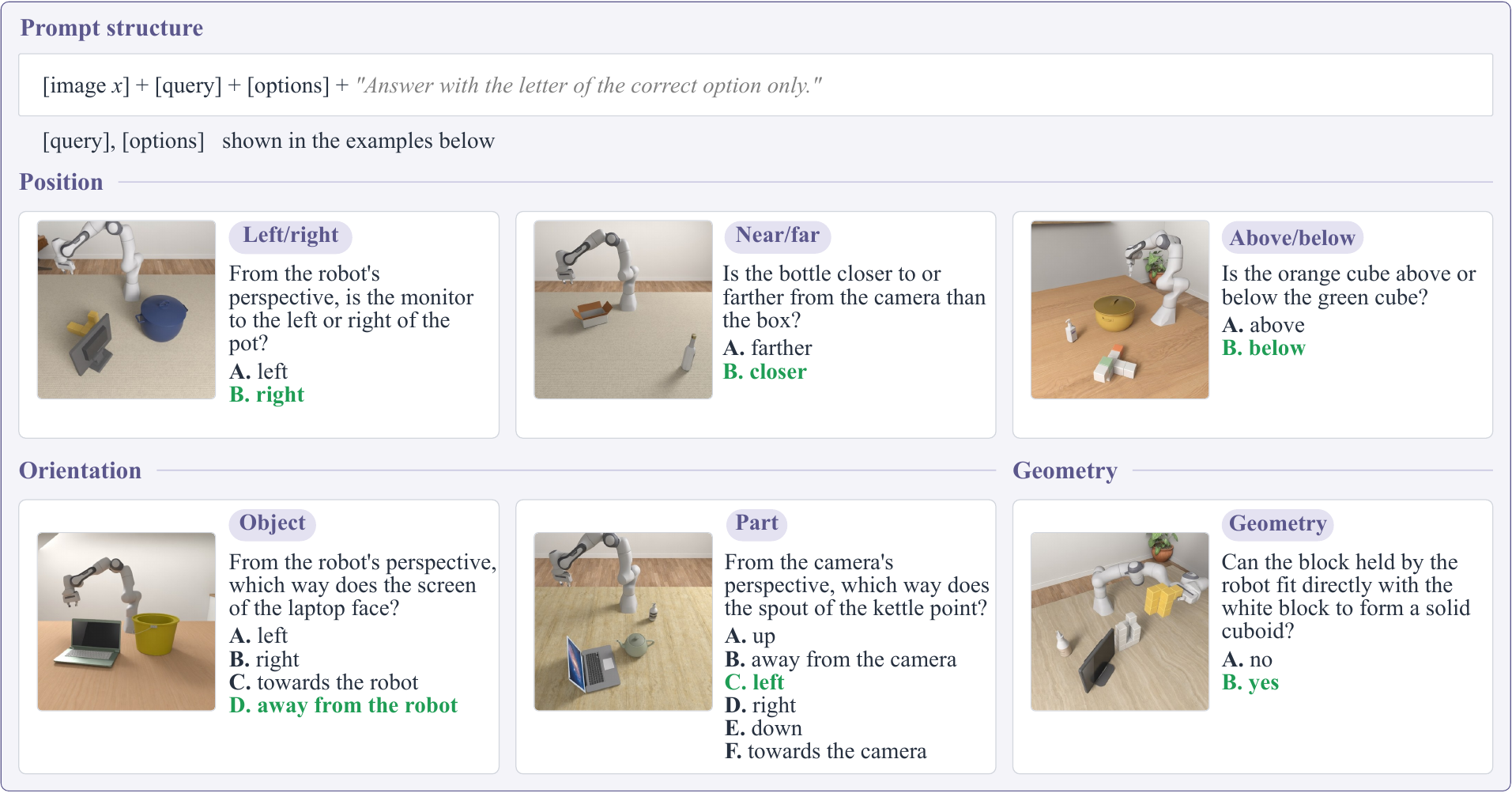}
  \caption{\textbf{Examples of \hlbox{constructionFill}{construction} questions.} One question per question type with its observation, query and options. The correct answer is in green. }
  \label{fig:suppl_examples_con}
\end{figure*}

\begin{figure*}[t]
  \centering
  \includegraphics[width=\textwidth]{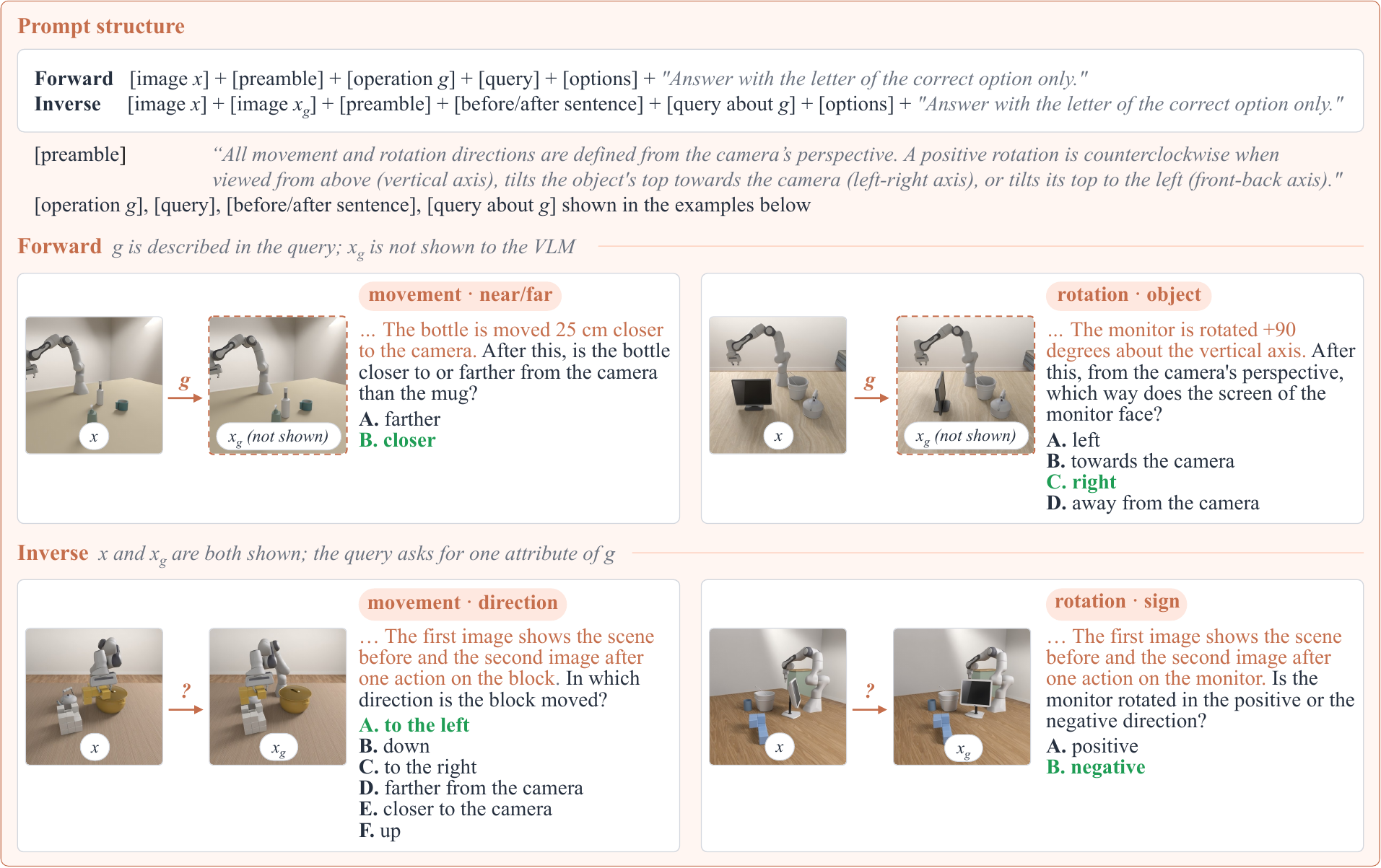}
  \caption{\textbf{Examples of \hlbox{operationFill}{operation} questions.} Forward questions describe the operation $g$ in the query, and inverse questions show the observations $x$ and $x_g$ before and after it. The correct answer is in green. }
  \label{fig:suppl_examples_op}
\end{figure*}

\begin{figure*}[t]
  \centering
  \includegraphics[width=\textwidth]{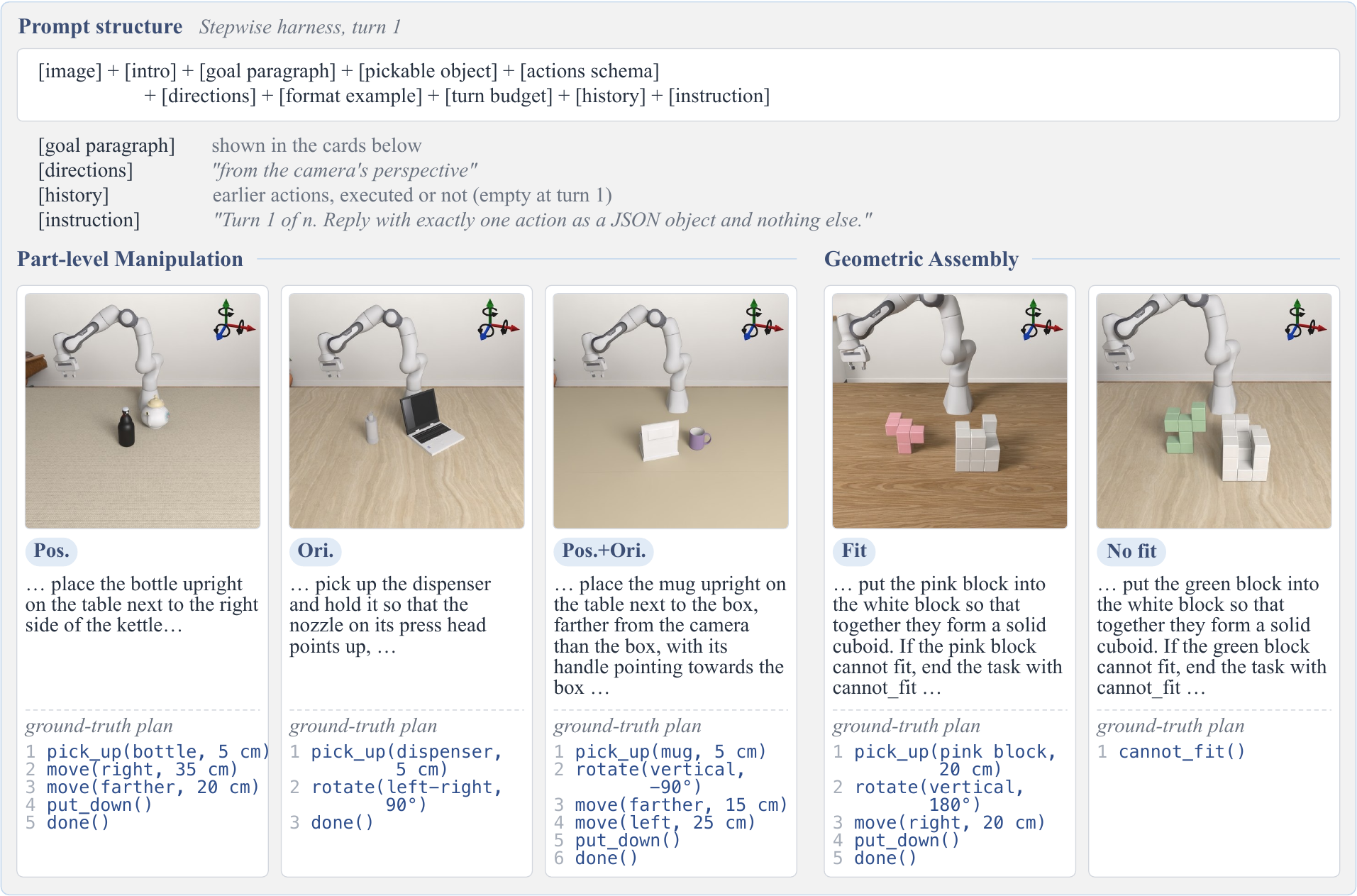}
  \caption{\textbf{Examples of the planning set.} One task per episode type with its first observation, goal and ground-truth plan. Abbreviations as in \cref{tab:notation}. }
  \label{fig:suppl_examples_task}
\end{figure*}

\Cref{fig:suppl_examples_con} shows one construction question of each question type, and \cref{fig:suppl_examples_op} shows forward and inverse operation questions.
\Cref{fig:suppl_examples_task} shows one task of each episode type of the planning set.
\section{Taxonomy of 3D representations in VLMs}
\label{sec:suppl_taxonomy}
As additional related work on 3D representations in VLMs, we further classify them into four kinds by their representation source and representation form (\cref{tab:taxonomy}).
The source is internal or external to the VLM, and the form is implicit or explicit.

\begin{table*}[t]
\caption{\textbf{Taxonomy of 3D representations in VLMs.} The rows give the representation source, and the columns give the representation form. We diagnose \hlbox{classBand}{Native} representations.}
\label{tab:taxonomy}
\centering
\footnotesize
\setlength{\tabcolsep}{4pt}
\setlength{\aboverulesep}{0pt}
\setlength{\belowrulesep}{0pt}
\renewcommand{\arraystretch}{1.25}
\begin{tabular}{@{\hspace{3pt}}>{\raggedright\arraybackslash}p{0.2\linewidth}>{\raggedright\arraybackslash}p{0.36\linewidth}>{\raggedright\arraybackslash}p{0.36\linewidth}@{\hspace{3pt}}}
\toprule
 & Implicit: 3D representations encoded in model activations
 & Explicit: 3D representations expressed in intermediate outputs \\
\midrule
Internal: 3D representations learned or generated by the VLM
 & \cellcolor{classBand}\textbf{Native}: the representation stays in the VLM's hidden states and can be strengthened by fine-tuning on spatial data~\cite{sat,robospatial} or by 3D supervision of visual tokens~\cite{3drs}
 & \textbf{Generated}: the VLM writes the representation out before it answers, as text with coordinates or
   maps~\cite{spatialreasoner,mindcube,spatialthinker}, as latent tokens~\cite{3dthinker,mirage,latentsketchpad,aurora}, or
   as generated views~\cite{wmvlm,mvot} \\
External: 3D representations provided by auxiliary models or tools
 & \textbf{Injected}: features computed by an auxiliary 3D model enter the VLM's activations, e.g.\ 3D
   foundation-model features~\cite{vgllm,spatialmllm,vlm3r,spatialstack} or point-cloud and 3D position
   features~\cite{3dllm}
 & \textbf{Tool-calling}: the VLM calls an auxiliary model or tool that produces an explicit spatial output, which the VLM then reads, e.g.\ 3D
   perception tools~\cite{spatialbot,apc,ego3dbench}, 3D reconstruction and rendering~\cite{gpt4scene}, or world
   models and view synthesis~\cite{mindjourney,spatialdreamer} \\
\bottomrule
\end{tabular}
\end{table*}

\section{Results on the diagnostic set}
\label{sec:suppl_diag_results}
This section gives the results on the diagnostic set, one table for each step of the cognitive reasoning chain: construction in \cref{tab:construction}, operation in \cref{tab:operation} and use in \cref{tab:use}.

\begin{table}[!t]
  \caption{\textbf{\hlbox{constructionFill}{Construction} per representation.} Normalized accuracy of the construction probe, averaged over the question types of each 3D property and the 18 VLMs. Abbreviations as in \cref{tab:notation}. Cells above chance (0) are \hlbox{aboveChance!40}{green} and cells below it are \hlbox{belowChance!40}{red}.}
  \label{tab:construction}
  \centering
  \footnotesize
  \setlength{\tabcolsep}{6pt}
  \setlength{\aboverulesep}{0pt}
  \setlength{\belowrulesep}{0pt}
  \renewcommand{\arraystretch}{1.15}
  \providecommand{\kof}[2]{\,{\tiny #1/#2}}
  \providecommand{\pts}[2]{\ifdim #1pt<0pt\relax\cellcolor{belowChance!\fpeval{round(min(1,-(#1)/100)*55)}}\else\cellcolor{aboveChance!\fpeval{round(min(1,(#1)/100)*55)}}\fi{#2}}%
  \begin{tabular}{@{\hspace{3pt}}l*{3}{c}@{\hspace{3pt}}}
  \toprule
  \multirow{2}{*}{Representation} & \multicolumn{3}{c}{\cellcolor{constructionFill}Construction} \\
    & Position & Orientation & Geometry \\
  \midrule
  Img. & \acc{87.7}{0} & \acc{48.6}{0} & \acc{19.4}{0} \\
  VObj. & \acc{93.9}{0} & \acc{58.1}{0} & \acc{19.6}{0} \\
  TObj. & \acc{86.1}{0} & \acc{52.4}{0} & \acc{4.9}{0} \\
  Ans. & \acc{84.5}{0} & \acc{54.0}{0} & \acc{4.3}{0} \\
  \bottomrule
  \end{tabular}
  \end{table}
  
\begin{table}[!t]
\caption{\textbf{\hlbox{operationFill}{Operation} per representation.} Forward (\cref{eq:fwd}) and inverse (\cref{eq:inv}) scores in points above the control, averaged over the VLMs. Abbreviations as in \cref{tab:notation}. Cells above the control (0) are \hlbox{aboveChance!40}{green} and cells below it are \hlbox{belowChance!40}{red}.}
\label{tab:operation}
\centering
\footnotesize
\setlength{\tabcolsep}{2.4pt}
\setlength{\aboverulesep}{0pt}
\setlength{\belowrulesep}{0pt}
\renewcommand{\arraystretch}{1.15}
\providecommand{\kof}[2]{\,{\tiny #1/#2}}
\providecommand{\pts}[2]{\ifdim #1pt<0pt\relax\cellcolor{belowChance!\fpeval{round(min(1,-(#1)/100)*55)}}\else\cellcolor{aboveChance!\fpeval{round(min(1,(#1)/100)*55)}}\fi{#2}}%
\begin{tabular}{@{\hspace{2pt}}l*{7}{c}@{\hspace{2pt}}}
\toprule
\multirow{2}{*}{Representation} & \multicolumn{2}{c}{\cellcolor{operationFill}Forward} & \multicolumn{5}{c}{\cellcolor{operationFill}Inverse} \\
  & Mov. & Rot. & Direction & Distance & Axis & Sign & Angle \\
\midrule
Img. & -- & -- & \pts{43.3235}{+43.3} & \pts{-1.0588}{$-$1.1} & \pts{0.5882}{+0.6} & \pts{0.7843}{+0.8} & \pts{0.3824}{+0.4} \\
VObj. & \pts{0.3136}{+0.3} & \pts{0.1480}{+0.1} & \pts{44.1471}{+44.1} & \pts{26.5882}{+26.6} & \pts{4.3529}{+4.4} & \pts{2.3137}{+2.3} & \pts{10.7647}{+10.8} \\
TObj. & \pts{12.1995}{+12.2} & \pts{0.5858}{+0.6} & \pts{15.1176}{+15.1} & \pts{28.6176}{+28.6} & \pts{4.1176}{+4.1} & \pts{-0.4706}{$-$0.5} & \pts{11.1176}{+11.1} \\
Ans. & \pts{27.5685}{+27.6} & \pts{4.7056}{+4.7} & \pts{12.1471}{+12.1} & \pts{34.2941}{+34.3} & \pts{6.5882}{+6.6} & \pts{4.7843}{+4.8} & \pts{23.5588}{+23.6} \\
\bottomrule
\end{tabular}
\end{table}

\begin{table}[!t]
\caption{\textbf{\hlbox{useFill}{Use} per representation.} Sufficiency and necessity scores (\cref{sec:use}) on construction and operation questions, in points above the random edit of the same norm and averaged over the VLMs. Abbreviations as in \cref{tab:notation}. Cells above the random edit (0) are \hlbox{aboveChance!40}{green} and cells below it are \hlbox{belowChance!40}{red}.}
\label{tab:use}
\centering
\footnotesize
\setlength{\tabcolsep}{3pt}
\setlength{\aboverulesep}{0pt}
\setlength{\belowrulesep}{0pt}
\renewcommand{\arraystretch}{1.15}
\providecommand{\kof}[2]{\,{\tiny #1/#2}}
\providecommand{\pts}[2]{\ifdim #1pt<0pt\relax\cellcolor{belowChance!\fpeval{round(min(1,-(#1)/100)*55)}}\else\cellcolor{aboveChance!\fpeval{round(min(1,(#1)/100)*55)}}\fi{#2}}%
\begin{tabular}{@{\hspace{2pt}}l*{4}{c}@{\hspace{2pt}}}
\toprule
\multirow{2}{*}{Representation} & \multicolumn{2}{c}{\cellcolor{useFill}Sufficiency score} & \multicolumn{2}{c}{\cellcolor{useFill}Necessity score} \\
  & Construction & Operation & Construction & Operation \\
\midrule
Img. & \pts{10.5750}{+10.6} & \pts{8.4084}{+8.4} & \pts{1.5556}{+1.6} & \pts{-0.3611}{$-$0.4} \\
VObj. & \pts{10.7201}{+10.7} & \pts{8.6360}{+8.6} & \pts{2.0278}{+2.0} & \pts{-0.5278}{$-$0.5} \\
TObj. & \pts{7.8287}{+7.8} & \pts{3.1871}{+3.2} & \pts{0.2083}{+0.2} & \pts{0.3403}{+0.3} \\
Ans. & \pts{25.4162}{+25.4} & \pts{25.9845}{+26.0} & \pts{1.2361}{+1.2} & \pts{-0.7014}{$-$0.7} \\
\bottomrule
\end{tabular}
\end{table}

\vspace*{0pt}\newpage\vspace*{0pt}\newpage\vspace*{0pt}\newpage
\section{Probing}
\label{sec:suppl_probing}
We use the notation of \cref{sec:framework}.
\begin{figure*}[t]
\centering
\includegraphics[width=0.86\textwidth]{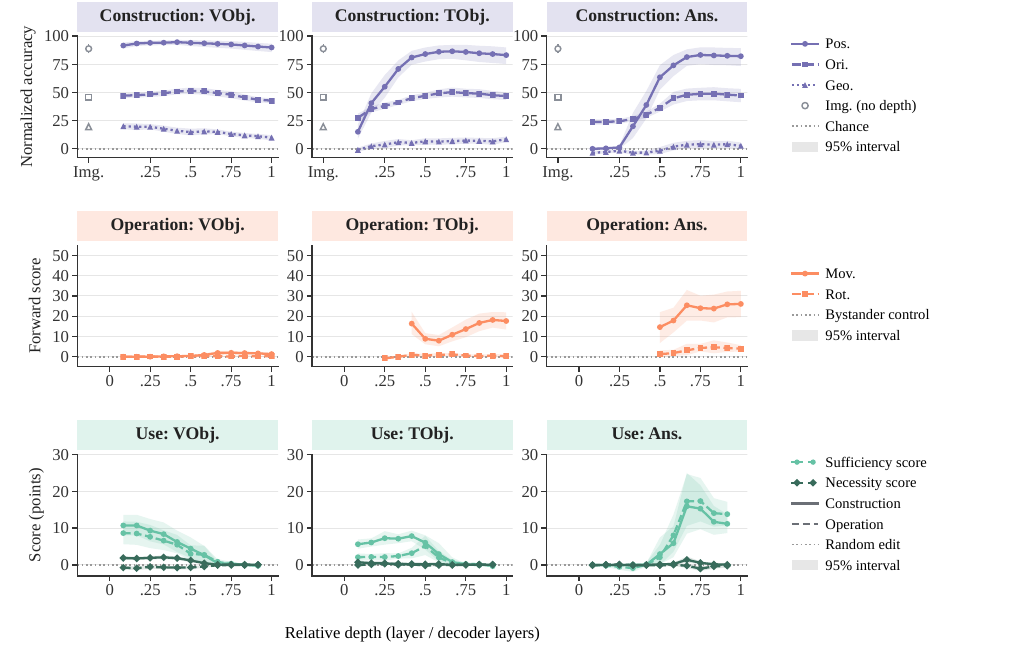}
\caption{\textbf{\hlbox{constructionFill}{Construction}, \hlbox{operationFill}{operation} and \hlbox{useFill}{use} along the depth of the language model.} Rows show the normalized accuracy of the construction probe in cross-validation on the training pool, the forward score and the use scores at the candidate layers, placed by relative depth. Img.\ is marked at the left of the construction row. In the use row, solid lines are on construction questions and dashed lines are on operation questions. Lines are averaged over the VLMs with 95\% intervals. Abbreviations as in \cref{tab:notation}. }
\label{fig:depth}
\end{figure*}

\subsection{Probe training}
\label{sec:suppl_probe_training}
Each VLM has one linear probe per representation and question type.
The training pool is built by the same rules as the test questions but from objects and polycube shapes that they never use.

\noindent\textbf{How each representation is read.}
Img.\ is read at the projector output, so its read layer is $\ell_p = 0$.
Img.\ and VObj.\ average the image tokens that the object covers, weighted by coverage.
At Img., VObj.\ and TObj., a probe takes the representation of the queried object or part.
For a relation, it takes the difference of the representations of the two objects or cubes, and for geometry, it concatenates the representations of the two polycubes.
Ans.\ is the last input token for every question type.
In the forward direction, the description of $g$ is placed before the image for VObj., so that the image tokens can attend to it.
In the inverse direction, Img.\ concatenates the embeddings of the two observations, since the vision encoder encodes them separately.
VObj.\ averages the object's tokens in $x_g$, which attend to $x$.
TObj.\ and Ans.\ are read as in construction.
LLaVA-1.5 takes one image per query, so the inverse direction is tested on the other 17 VLMs.
For each attribute of $g$, the single-observation probe reads the observation, $x$ or $x_g$, on which it is more accurate in object-grouped five-fold cross-validation on the training pool.

\subsection{Layer selection}
We read 12 evenly spaced candidate decoder layers of each language model.
For each of VObj., TObj.\ and Ans., the read layer of a VLM is the candidate layer at which its construction probes are most accurate.
This accuracy is measured in object-grouped five-fold cross-validation on the training pool and averaged over the question types.
The probes are then refit on the whole training pool at that layer and tested on the 1,062 test questions.
The bystander and inverse probes read the same layer.
In the use tests, Img.\ and Ans.\ are edited where they are read, at the projector output and at the read layer.
VObj.\ and TObj.\ are each edited at one candidate layer shared by all VLMs.
It is the layer where edits on construction questions change the answers most on development scenes, averaged over the VLMs.
\Cref{fig:depth} shows these readouts and scores at the candidate layers.

\subsection{Class separation weakens from position to orientation and geometry}
\label{sec:suppl_planes}
\begin{figure*}[t]
\centering
\includegraphics[width=\textwidth]{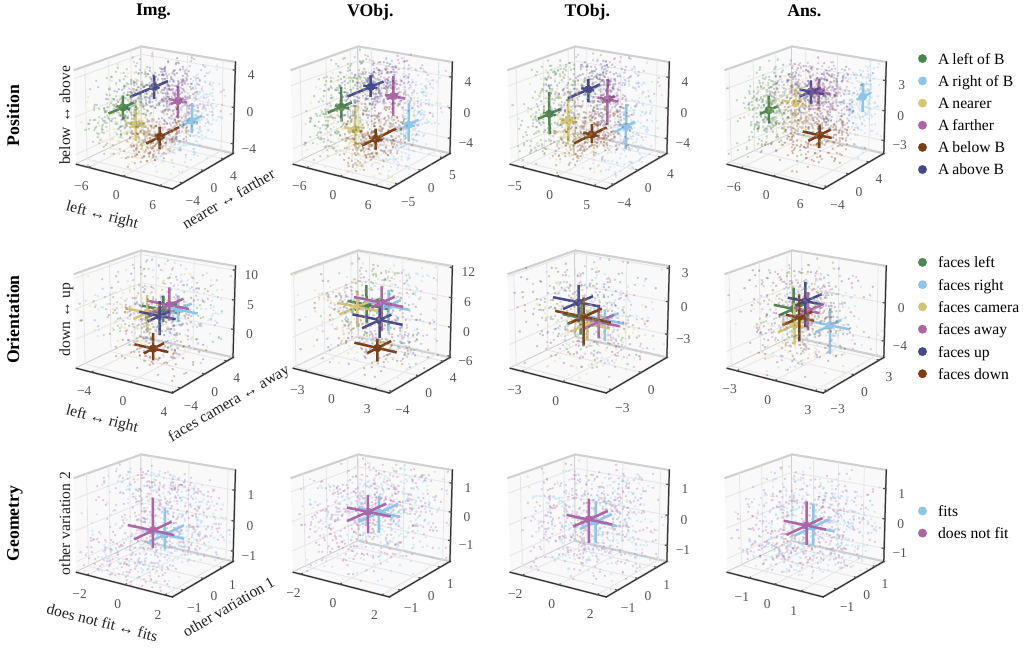}
\caption{\textbf{Class separation at \hlbox{constructionFill}{construction} in 3D visualization.} Hidden states of Qwen2.5-VL-7B at the four representations, projected onto three discriminant axes fitted on other objects. Large markers are class medians with interquartile crosses. Abbreviations as in \cref{tab:notation}. }
\label{fig:suppl_planes3d}
\end{figure*}

\begin{figure*}[p]
\centering
\includegraphics[width=\textwidth]{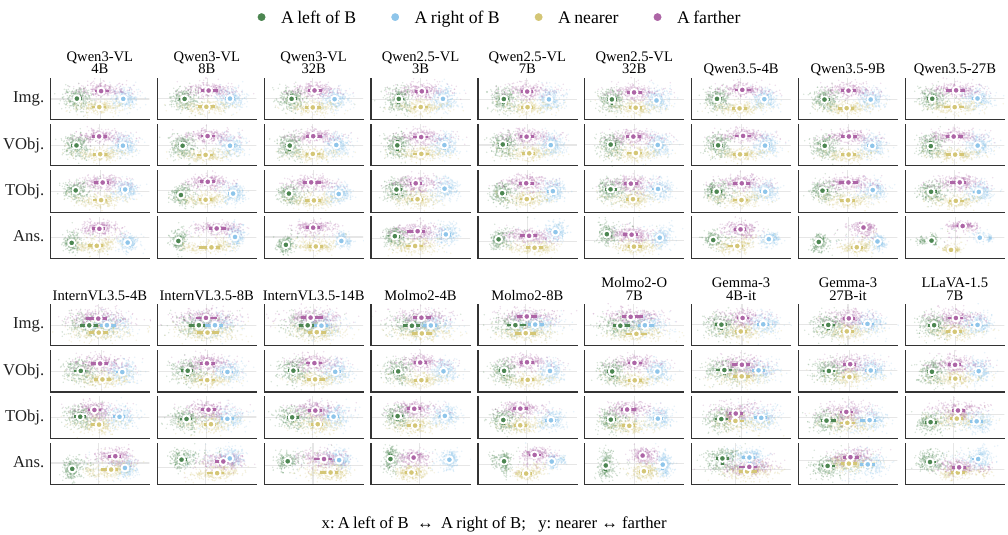}
\caption{\textbf{Class separation of position at \hlbox{constructionFill}{construction} for the 18 VLMs.} One panel per VLM and representation, projected onto two discriminant axes fitted on other objects. Large markers are class medians with interquartile crosses. Abbreviations as in \cref{tab:notation}. }
\label{fig:suppl_geo_pos}
\end{figure*}

\begin{figure*}[p]
\centering
\includegraphics[width=\textwidth]{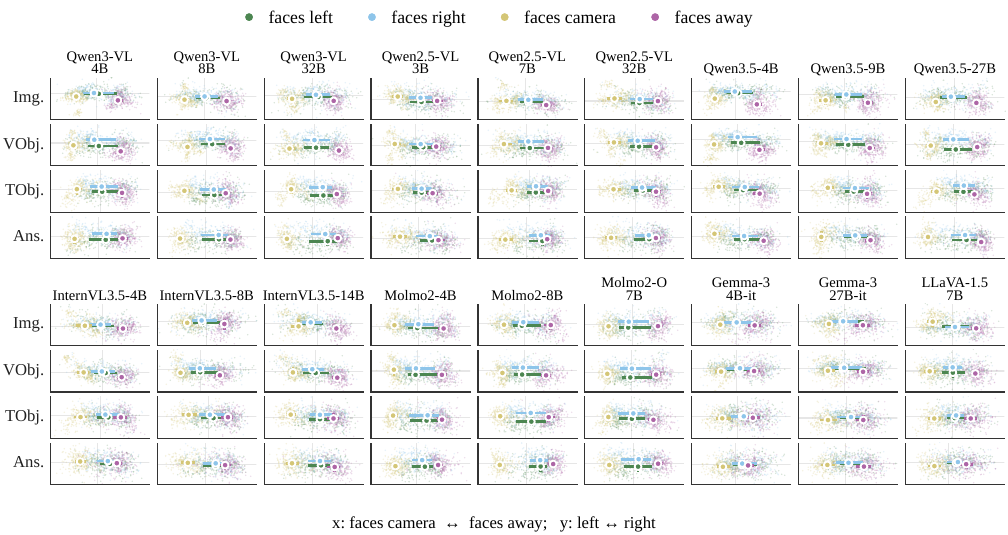}
\caption{\textbf{Class separation of orientation at \hlbox{constructionFill}{construction} for the 18 VLMs.} One panel per VLM and representation, projected onto two discriminant axes fitted on other objects. Large markers are class medians with interquartile crosses. Abbreviations as in \cref{tab:notation}. }
\label{fig:suppl_geo_obj}
\end{figure*}

\begin{figure*}[p]
\centering
\includegraphics[width=\textwidth]{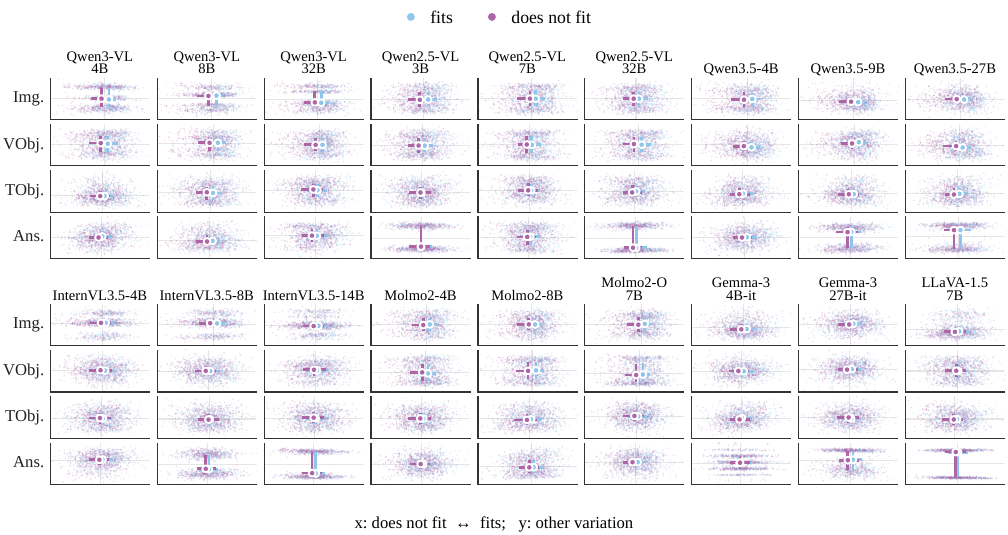}
\caption{\textbf{Class separation of geometry at \hlbox{constructionFill}{construction} for the 18 VLMs.} One panel per VLM and representation, projected onto two discriminant axes fitted on other objects. Large markers are class medians with interquartile crosses. Abbreviations as in \cref{tab:notation}. }
\label{fig:suppl_geo_geo}
\end{figure*}

\Cref{fig:planes} shows class separation with \mbox{Qwen2.5-VL-7B} as an example, and \cref{fig:suppl_geo_pos,fig:suppl_geo_obj,fig:suppl_geo_geo} show it for all 18 VLMs.
In all 18 VLMs, class separation weakens from position to orientation and geometry, and the two classes of geometry largely overlap at all four representations.
The same pattern holds in 3D visualization, shown in \cref{fig:suppl_planes3d} with \mbox{Qwen2.5-VL-7B} as an example.
Each panel projects the hidden states at one representation onto discriminant axes fitted on other objects.

\subsection{MLP probes reproduce the linear-probe findings}
\label{sec:suppl_mlp}
\begin{figure*}[t]
\centering
\includegraphics[width=\textwidth]{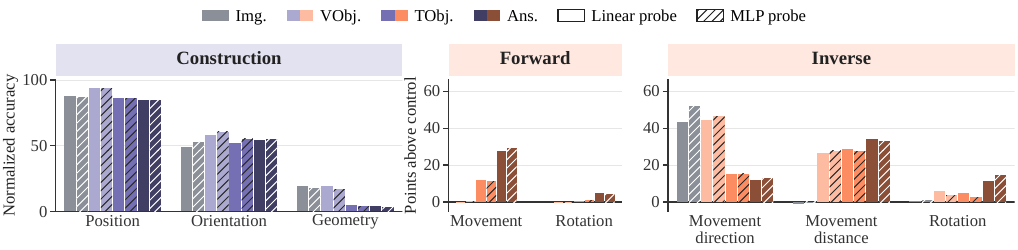}
\caption{\textbf{Linear and MLP probes.} The \hlbox{constructionFill}{construction} panel shows normalized accuracy per 3D property, and the \hlbox{operationFill}{operation} panels show forward and inverse scores. Bars are averaged over the VLMs, solid for the linear probe and hatched for the MLP probe. Abbreviations as in \cref{tab:notation}. }
\label{fig:mlp}
\end{figure*}

A linear probe can miss information that a representation encodes nonlinearly.
Such a miss would matter for the two main findings that report an absence: geometry is read only slightly above chance, and a described rotation updates no representation.
The readouts of construction and operation come from the construction, bystander and inverse probes.
To test whether these absences come from their linearity, we refit each of them as a multilayer perceptron (MLP) probe with one hidden layer of rectified linear units.
Each MLP probe reads the same representation and is trained on the same training pool.
As shown in \cref{fig:mlp}, the MLP probes reproduce the linear-probe findings.
They read geometry no better, and the forward score of a described rotation stays near zero.

\subsection{Hardware}
\label{sec:suppl_compute}
We extract hidden states and run the edits on NVIDIA H100 and A100 GPUs, and we fit the probes on CPUs.
\section{Answer-level accuracy on the diagnostic set}
\label{sec:suppl_answer_level}
\subsection{The answers express only part of the readout}
Each VLM sees the image and the multiple-choice query and answers with one option letter under greedy decoding.
Averaged over the question types and the 18 VLMs, the answers reach a normalized accuracy of 51 for position and 15 for orientation and stay at chance for geometry.
For position and orientation, these are 60\% and 27\% of the readout at Ans.

\subsection{The answers lean on 2D image cues rather than on the 3D relation}
\label{sec:suppl_cue}
\begin{figure}[t]
\centering
\includegraphics[width=\columnwidth]{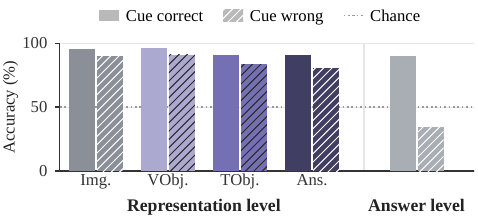}
\caption{\textbf{\hlbox{constructionFill}{Construction} readout and answers on above/below questions, split by the 2D image cue.} The cue takes the cube higher in the image as above. Bars are averaged over the 18 VLMs, solid where the cue gives the correct answer and hatched where it gives a wrong one. Abbreviations as in \cref{tab:notation}. }
\label{fig:cue}
\end{figure}
On above/below questions, the 2D image cue takes the cube that appears higher in the image as above.
As shown in \cref{fig:cue}, the answers are right on 90.0\% of the questions where this cue gives the correct answer and on only 34.8\% where it does not.
The readout of every representation stays at 81 to 96\% in both cases.

\subsection{The answers treat a rotation as a flip to the opposite orientation}
\label{sec:suppl_flip}
\begin{figure}[t]
\centering
\includegraphics[width=\columnwidth]{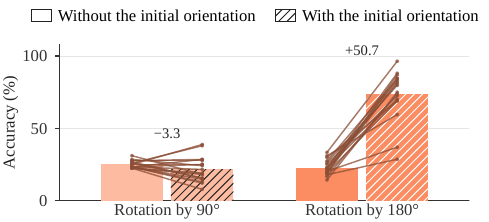}
\caption{\textbf{Accuracy with the initial orientation given.} Forward questions about object and part orientation, without (solid) and with (hatched) one added sentence that gives the initial orientation. Bars are averaged over the 18 VLMs, and lines are single VLMs. }
\label{fig:flip}
\end{figure}

A forward question describes a rotation of an object from its initial orientation, and its answer is the orientation after the rotation.
A wrong answer can come from misreading the initial orientation or from failing to apply the rotation.
To separate the two, we add one sentence to the query that gives the initial orientation in the words of the options.
The sentence comes right before the description of the rotation, \eg, ``Initially, from the camera's perspective, the screen of the laptop faces left.''
The image and the rest of the query stay unchanged.
Every VLM answers the same forward questions about object and part orientation with and without this sentence.
As shown in \cref{fig:flip}, adding this sentence changes accuracy by $-3.3$ points on rotations by 90\textdegree{} but raises it by 50.7 points on rotations by 180\textdegree{}, where the opposite orientation is correct.
This rise holds in all 18 VLMs.
On rotations by 90\textdegree{}, 48.6\% of the answers with the sentence choose the opposite of the initial orientation.
Thus, the answers fail to apply the rotation rather than to read the initial orientation.

\begin{figure*}[t]
  \centering
  \includegraphics[width=\textwidth]{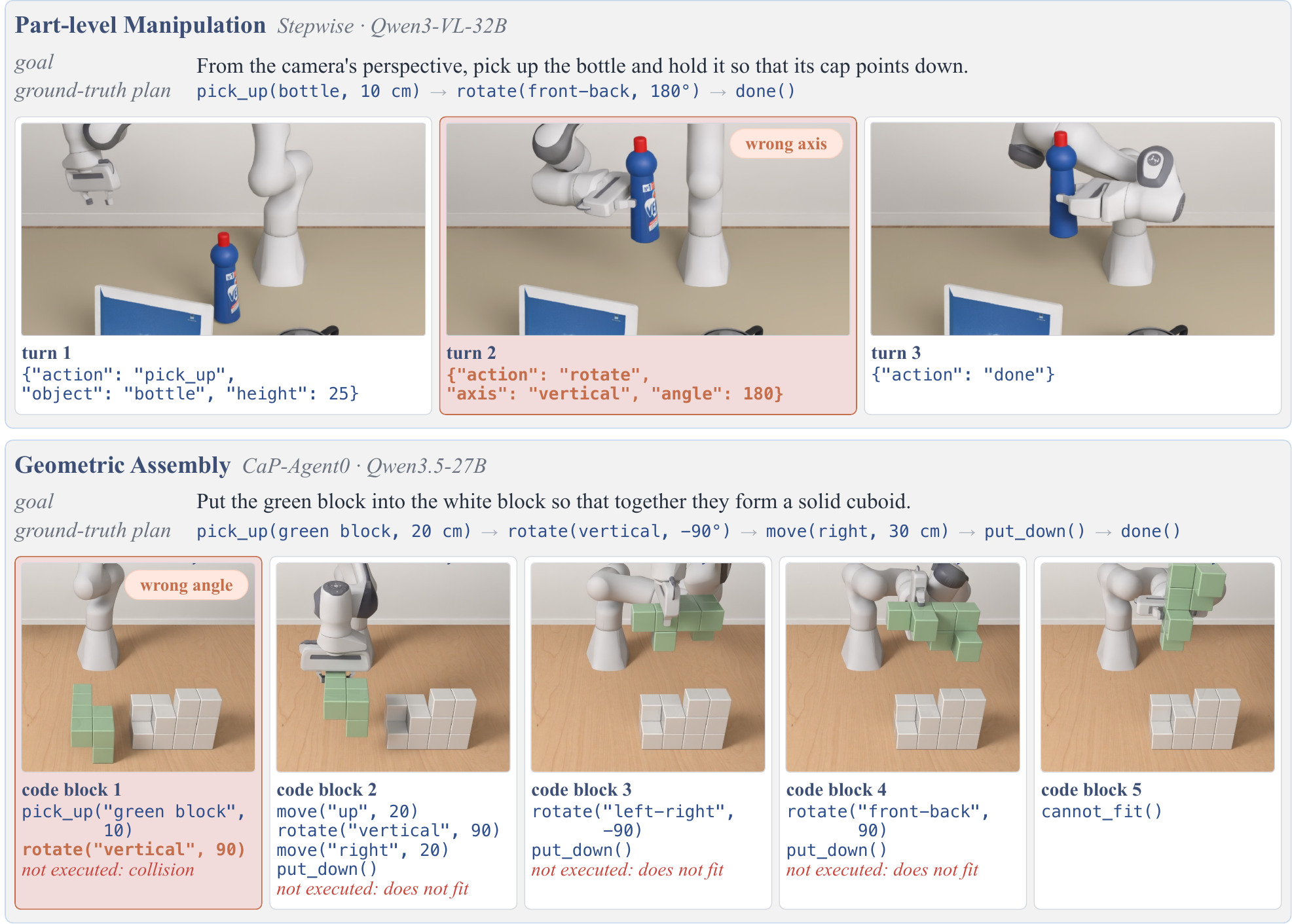}
  \caption{\textbf{Failed episodes on the planning set.} Top: Part-level Manipulation with Stepwise (Qwen3-VL-32B). Bottom: Geometric Assembly with CaP-Agent0 (Qwen3.5-27B). Each card shows an observation and the actions the VLM takes after it, and the decisive error is highlighted with its type. }
  \label{fig:suppl_case_study}
\end{figure*}
\section{Case studies}
\label{sec:suppl_cases}

\Cref{fig:suppl_case_study} shows one failed episode of each task type of the planning set.
In both, the decisive error (\cref{sec:suppl_failures}) is a wrong rotation.
In Part-level Manipulation, \mbox{Qwen3-VL-32B} with Stepwise rotates the bottle by 180° about the vertical axis instead of the front-back axis, a wrong axis.
In Geometric Assembly, Qwen3.5-27B with CaP-Agent0 rotates the polycube by 90° instead of $-$90° about the vertical axis, a wrong angle.
\section{Closed-loop evaluation on the planning set}
\label{sec:suppl_task}

\subsection{Harnesses}
\label{sec:suppl_harness}
\begin{figure*}[t]
  \centering
  \includegraphics[width=0.495\textwidth]{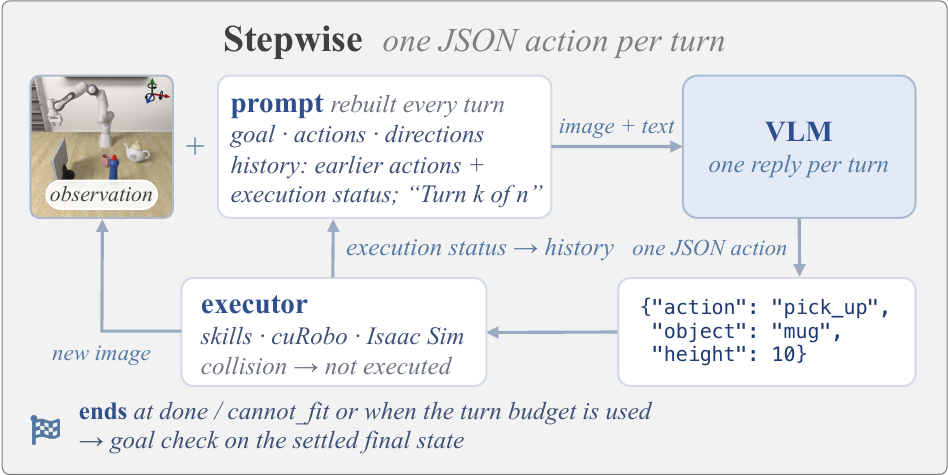}\hfill
  \includegraphics[width=0.495\textwidth]{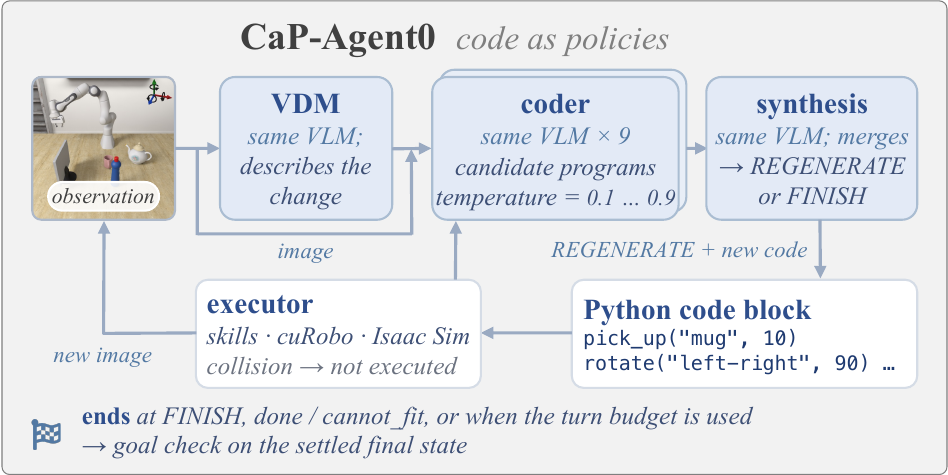}
  \caption{\textbf{The two harnesses on the planning set.} Left: Stepwise, one JSON action per turn. Right: CaP-Agent0~\cite{capx}, in which the same VLM describes the observations as the visual differencing model (VDM), writes candidate programs and merges them into one code block. }
  \label{fig:suppl_harness}
\end{figure*}

\Cref{fig:suppl_harness} shows the two harnesses.
Both harnesses start every episode from the initial scene with the robot arm at its home pose.
Each observation shows the VLM one image from a fixed camera.

\noindent\textbf{Actions.}
\textit{pick\_up(object, height)} grasps an object and lifts it until its bottom is 5 to 30\,cm above the table.
\textit{move(direction, distance)} moves the held object by 5 to 40\,cm to the left, to the right, up, down, closer to the camera or farther from it.
\textit{rotate(axis, angle)} rotates the held object about the vertical, the left-right or the front-back axis by $\pm$30°, $\pm$60°, $\pm$90° or 180°.
The prompt states the positive direction of each axis, and an icon in the image shows the axes.
\textit{put\_down()} lowers the held object and releases it, and in Geometric Assembly it inserts the polycube only when the two polycubes fit together directly.
\textit{done()} ends the episode, and \textit{cannot\_fit()} ends a Geometric Assembly episode with the judgement that the polycube cannot fit.
Actions are executed in Isaac Sim on NVIDIA L40S and RTX GPUs with motion planned by cuRobo~\cite{curobo}.
An action that would cause a collision is not executed and leaves the scene unchanged.

\noindent\textbf{Turn budget and success.}
Every action call counts as one turn.
The turn budget grows with the ground-truth plan.
A No fit task takes the turn budget of the plan that would insert its polycube if it fitted, so that the budget does not reveal whether the polycube fits.
Success is judged on the final state after the physics settles.
In Part-level Manipulation, a goal on position requires the object upright on the table on the named side of the reference object.
A goal on orientation requires the named part of the held object to point in the named direction.
A goal on both requires both conditions, with the part pointing at the reference object.
In Geometric Assembly, Fit requires the two polycubes to form a solid cuboid, and No fit requires \textit{cannot\_fit()}.
Each task is confirmed to be feasible by executing its ground-truth plan successfully in repeated simulation runs.

\noindent\textbf{Stepwise.}
At each turn, the VLM receives the current image and a prompt and replies with one action as a JSON object.
The prompt holds the goal, the actions, the directions from the camera's perspective, the turn budget and every earlier action with whether it was executed.

\noindent\textbf{CaP-Agent0.}
We keep the agent of CaP-X~\cite{capx}, its prompts and its loop, and replace its robot interface with the actions above.
The VLM under test plays every role.
It describes the first observation and the change after each code block, writes nine candidate programs and merges them into one code block.
A call that is not executed raises an error, which reaches the VLM with the next observation.
After each code block, the VLM either finishes the episode or writes a new code block.

\noindent\textbf{Decoding.}
In Stepwise, the open-source VLMs decode greedily, and GPT-6 Astra and Gemini 3.1 Pro use the default decoding of their providers.
CaP-Agent0 keeps the sampling settings of CaP-X.

\subsection{Open-source VLMs mostly fail the planning set}
\begin{figure*}[t]
  \centering
  \includegraphics[width=\textwidth]{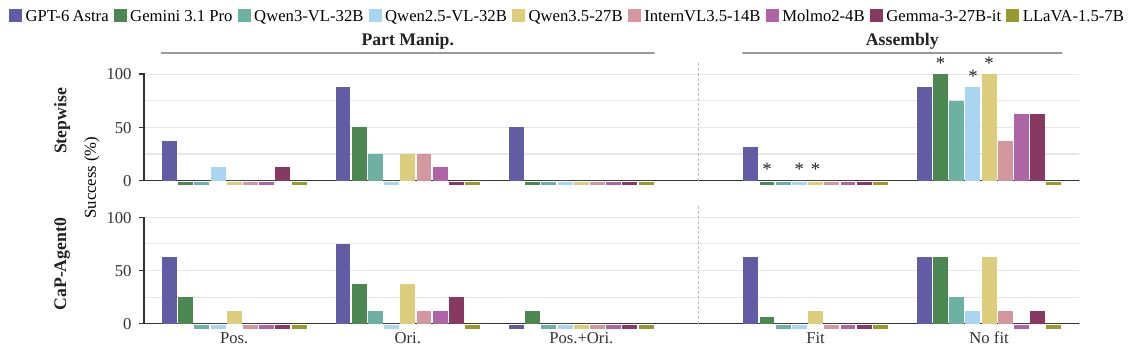}
  \caption{\textbf{Success per episode type on the planning set.} GPT-6 Astra, Gemini 3.1 Pro and the most successful open-source VLM of each series, on the 48 tasks that all VLMs run with both harnesses. A short bar below the axis marks a success of 0, and * marks a VLM that ends at least 90\% of its Geometric Assembly episodes with \textit{cannot\_fit()} in that harness. Abbreviations as in \cref{tab:notation}. }
  \label{fig:suppl_task_success}
\end{figure*}
\Cref{fig:suppl_task_success} gives the success per episode type on the 48 tasks (8 per Part-level Manipulation episode type, 16 Fit and 8 No fit) that all VLMs run with both harnesses.
On these tasks, the open-source VLMs succeed in only 9.4\% with Stepwise and 4.7\% with CaP-Agent0 on average, and at most 22.9\% (Qwen3.5-27B), whereas GPT-6 Astra and Gemini 3.1 Pro succeed in 54.2\% and 25.0\% with both harnesses.

\subsection{Failure analysis}
\label{sec:suppl_failures}
\noindent\textbf{Decisive errors.}
To attribute each failed episode to one action, we find its decisive error with fixed rules on the recorded states.
We judge each movement or rotation by the effect it would have on the state at its turn, whether or not it is executed.
It is an error when this effect does not bring the object closer to the goal while another action would, or when the action is not executed.
The decisive error is the first error that is never recovered: no later state is closer to the goal than the state before it.
Without such an error, the end of the episode decides: a \textit{done()} before the goal is met, a wrong fit judgement or a used-up budget.
In \cref{fig:task_failures}, a wrong rotation or movement picks a wrong axis, angle or distance, including the opposite way.
A wrong judgement stops before the goal, acts where nothing was needed or misjudges whether the polycube fits.
A collision is an otherwise correct action that is not executed because it would collide.
Other errors are a wrong object, an invalid reply, an unreachable action and a used-up budget.

\noindent\textbf{Probes on the Stepwise turns.}
A decisive rotation error can come from misreading the direction of the part named in the goal before the action (the state) or from failing to represent its direction after the rotation (the outcome).
To separate the two, we feed each recorded turn of a Part-level Manipulation episode back to the VLM with its reply and read its hidden states at the read layer of Ans.
One linear probe reads the state at Ans., and another reads the outcome at the last token of the action.
We compare the decisive rotation errors with the correct rotations of the same VLM that start from the same direction.
At the decisive errors, the state before the action is read at least as well as at the matched correct turns (42.3\% versus 32.6\%).
The outcome of a rotation, however, is read no better than from the action text or the state before the action alone in any of the 18 VLMs, as for a described rotation (\cref{fig:depth}).

\end{document}